\documentclass{article}
\usepackage[T1]{fontenc}
\usepackage{natbib}
\usepackage{fullpage}
\usepackage{amsmath,amsthm,amssymb,amsfonts,bbm,bm}
\usepackage{algorithm}
\usepackage{algorithmic}
\usepackage{enumitem}
\usepackage{etoolbox}
\usepackage{graphicx}
\usepackage{hyperref}
\usepackage{subfigure}
\usepackage[small,bf]{caption}
\usepackage{multirow}
\newcommand\EnumPrefix{}
\usepackage{xcolor}

\newlist{senenum}{enumerate}{10}
\setlist[senenum]{label=\arabic*.,ref=\EnumPrefix,leftmargin=*}

\AtBeginEnvironment{lemmalist}{\renewcommand\EnumPrefix{\thelemmalist.\arabic*}}

\newtheorem{theorem}{Theorem}[section]
\newtheorem{lemma}{Lemma}[section]
\newtheorem{lemmalist}{Lemma}[section]
\newtheorem{definition}{Definition}[section]

\newtheorem{proposition}{Proposition}[section]
\newtheorem{assumption}{Assumption}[section]
\numberwithin{equation}{section}

\newcommand{\bx}{{\boldsymbol{x}}}
\newcommand{\bX}{{\boldsymbol{X}}}
\newcommand{\cS}{{\mathcal{S}}}

\newcommand{\minitab}[3][l]{\begin{tabular}{#1}#2\end{tabular}}

\newcommand{\E}{\operatorname{\mathbb{E}}}

\hypersetup{
  colorlinks,
  citecolor=blue,
  linkcolor=red,
  urlcolor=blue}

\title{Adaptive Smoothness-weighted Adversarial Training for Multiple Perturbations with Its Stability Analysis}

\author{%
  Jiancong Xiao$^1$, Zeyu Qin$^1$, Yanbo Fan$^{2,}\thanks{Correspondence Authors.}$\ ,\\ Baoyuan Wu$^{1,3}$, Jue Wang$^{2}$, Zhi-Quan Luo$^{1,3,*}$\\
   $^1$The Chinese University of Hong Kong, Shenzhen; \\ 
   $^2$Tencent AI Lab; $^3$Shenzhen Research Institute of Big Data\\
  \texttt{jiancongxiao@link.cuhk.edu.cn, zeyuqin@cuhk.link.edu.cn,} \\
    \texttt{fanyanbo0124@gmail.com,wubaoyuan@cuhk.edu.cn,} \\
  \texttt{arphid@gmail.com, luozq@cuhk.edu.cn}
}
\date{}

\begin{document}

\maketitle

\begin{abstract}
 Adversarial Training (AT) has been demonstrated as one of the most effective methods against adversarial examples. While most existing works focus on AT with a single type of perturbation (\emph{e.g.,} the $\ell_\infty$ attacks), DNNs are facing threats from different types of adversarial examples. Therefore, adversarial training for multiple perturbations (ATMP) is proposed to generalize the adversarial robustness over different perturbation types (in $\ell_1$, $\ell_2$, and $\ell_\infty$ norm-bounded perturbations). However, the resulting model exhibits trade-off between different attacks. Meanwhile, there is no theoretical analysis of ATMP, limiting its further development. In this paper, we first provide the smoothness analysis of ATMP and show that $\ell_1$, $\ell_2$, and $\ell_\infty$ adversaries give different contributions to the smoothness of the loss function of ATMP. Based on this, we develop the stability-based excess risk bounds and propose adaptive smoothness-weighted adversarial training for multiple perturbations. Theoretically, our algorithm yields better bounds. Empirically, our experiments on CIFAR10 and CIFAR100 achieve the state-of-the-art performance against the mixture of multiple perturbations attacks.
\end{abstract}
\section{Introduction}
\begin{figure*}[htbp]
\vskip -0.15in
\centering
\scalebox{0.96}{
\subfigure[]{
\begin{minipage}[htp]{0.24\linewidth}
\centering
\includegraphics[width=1.2in]{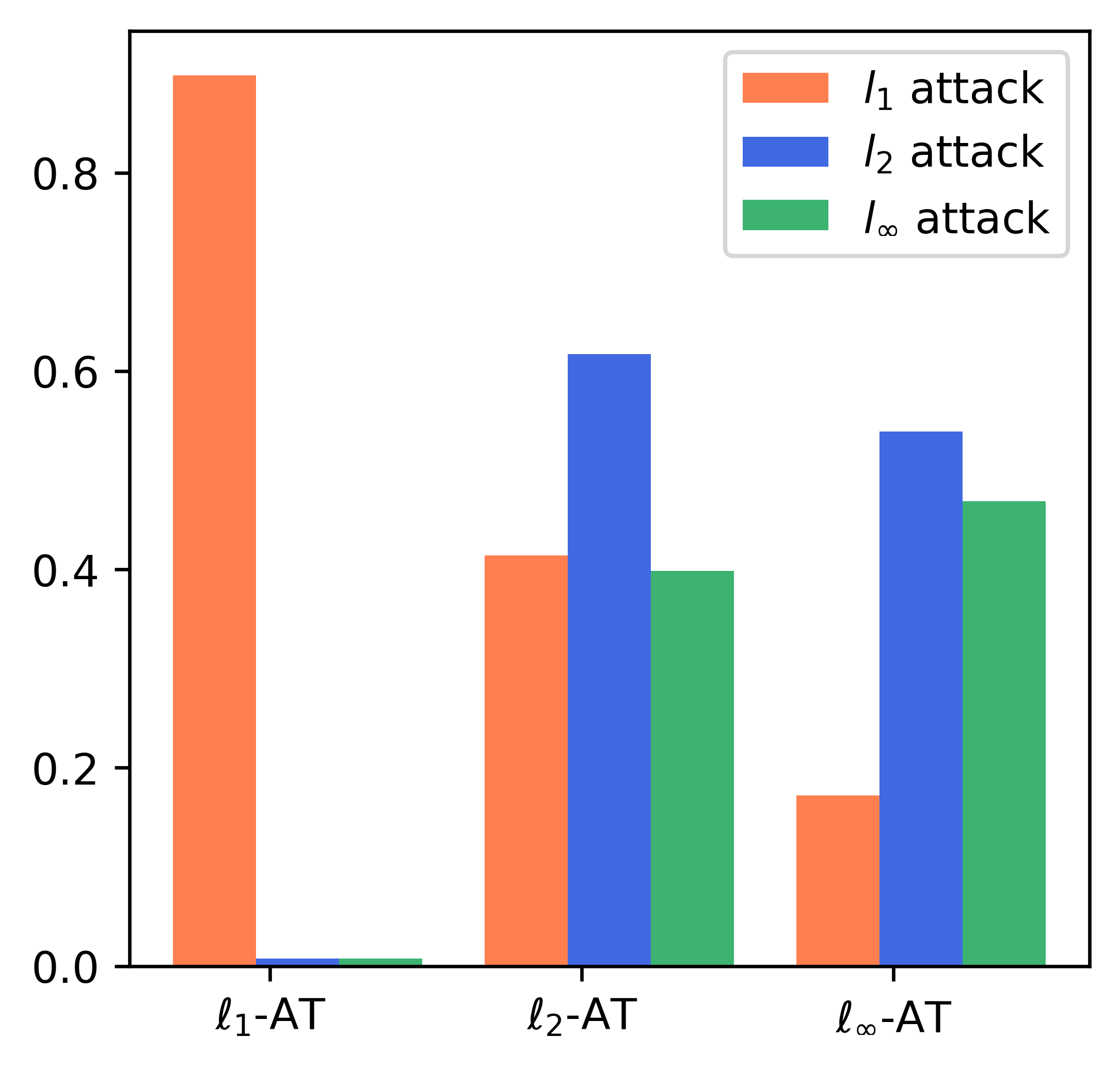}
\end{minipage}%
}%

\subfigure[]{
\begin{minipage}[htp]{0.24\linewidth}
\centering
\includegraphics[width=1.2in]{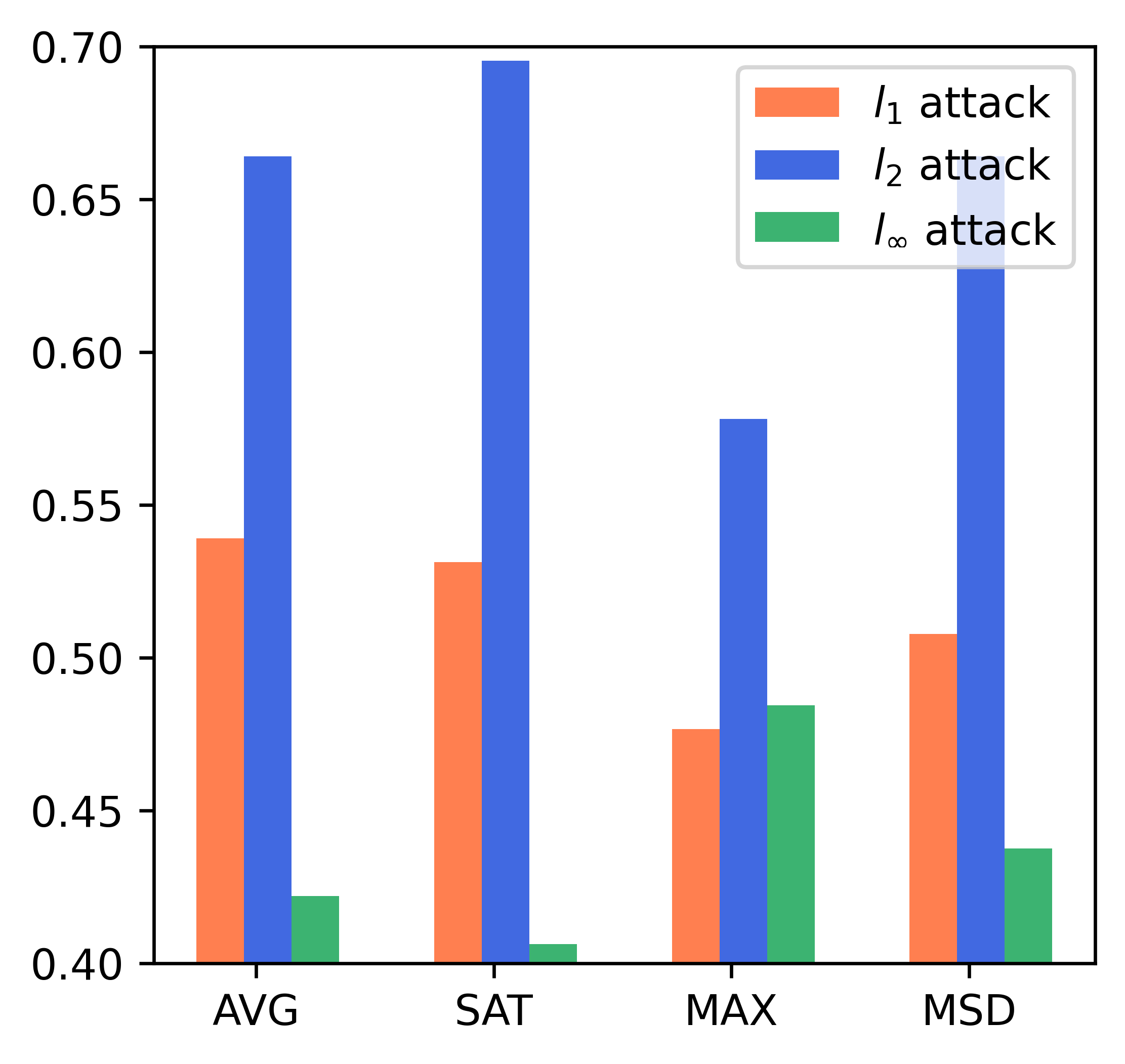}
\end{minipage}%
}%

\subfigure[]{
\begin{minipage}[htp]{0.24\linewidth}
\centering
\includegraphics[width=1.2in]{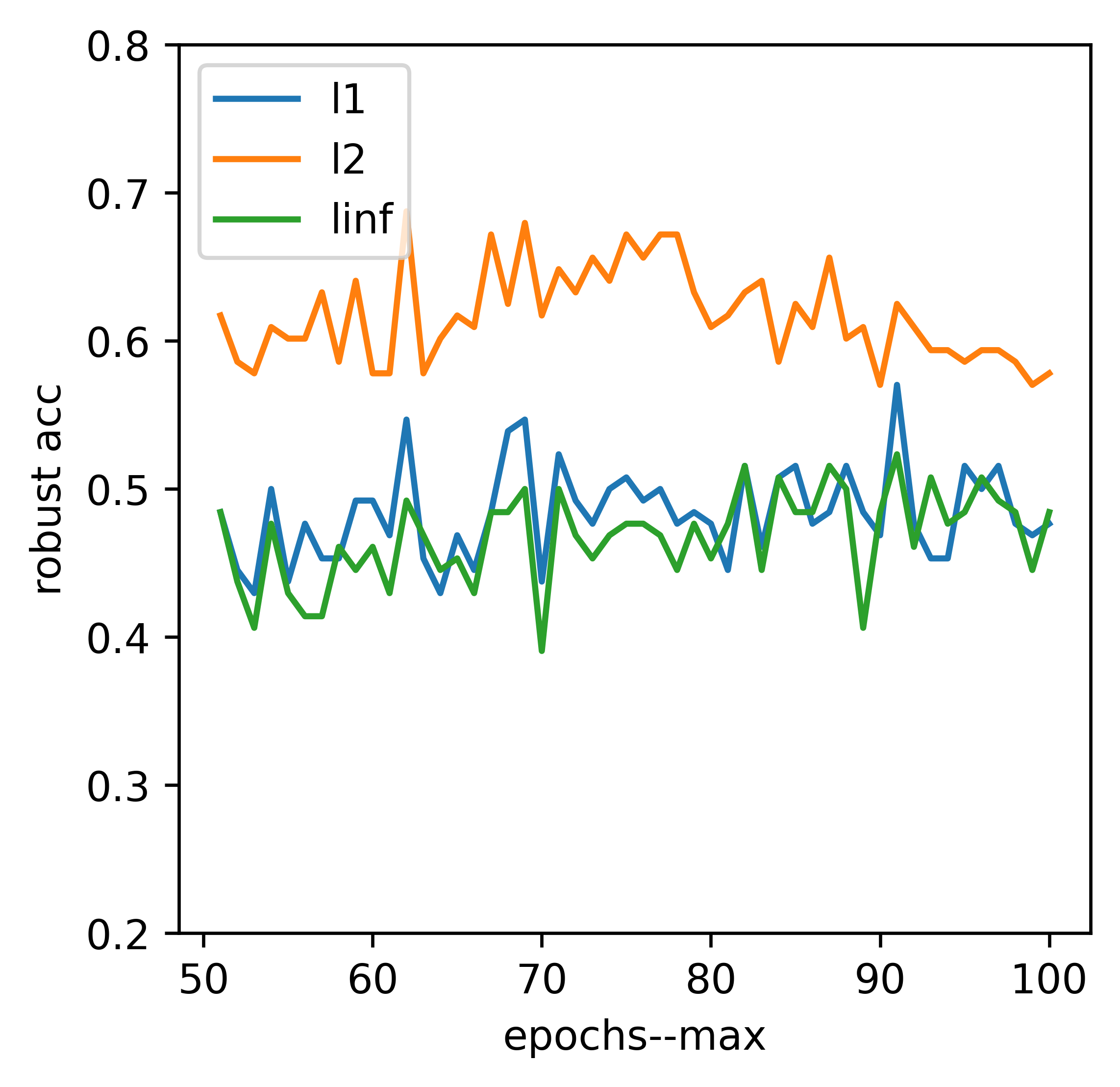}
\end{minipage}
}%

\subfigure[]{
\begin{minipage}[htp]{0.24\linewidth}
\centering
\includegraphics[width=1.2in]{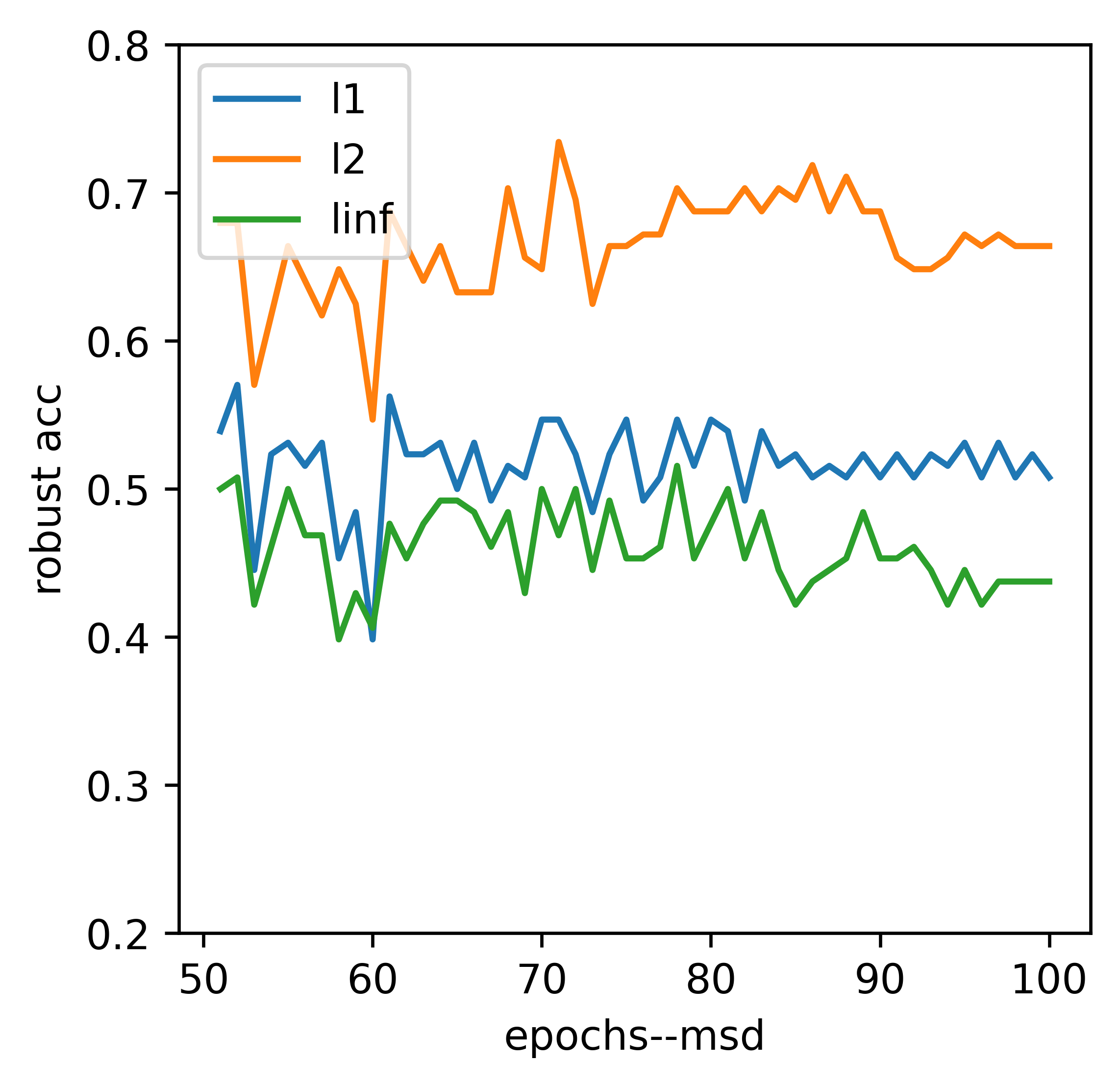}
\end{minipage}
}}

\centering
\vskip -0.15in
\caption{Crucial issues of adversarial training for multiple perturbations. (a) Performance of adversarial training with a single type perturbation against other type attacks. (b) Trade-off between different types of adversaries of four algorithms for ATMP. (c) Robust test accuracy fluctuate between different epochs using MAX. (d) Robust test accuracy fluctuate between different epochs using MSD.
}
\label{fig-square}
\vskip -0.15in
\end{figure*}

Deep neural networks (DNNs) are shown to be vulnerable to adversarial examples \citep{goodfellow2014explaining,szegedy2013intriguing}, where a small and malicious perturbation can cause incorrect predictions. Adversarial training (AT) \citep{madry2017towards}, which augments training data with $\ell_p$ norm-bounded adversarial examples, is one of the most effective methods to increase the robustness of DNNs against adversarial attacks. Currently, most existing works focus on adversarial training with a single type of attack, \emph{e.g.,} the $\ell_\infty$ attack \citep{raghunathan2019adversarial,gowal2020uncovering}. However, some recent works \citep{tramer2019adversarial} have experimentally demonstrated that the DNNs trained with a single type of adversarial attack cannot provide well defense against other types of adversarial examples. Fig. \ref{fig-square} (a) provides an example on CIFAR-10. The plot shows that the $\ell_1$ adversarial training cannot defend the $\ell_2$ and $\ell_\infty$ attacks (its robust accuracy are 0\%). The $\ell_\infty$ adversarial training can provide some extent of defense against $\ell_1$ and $\ell_2$ attacks (with accuracy being 17.19\% and 53.91\%). But it is not comparable to the performance of $\ell_1$ and $\ell_2$ adversarial training, 89.84\%, and 61.72\%, respectively.

For better robustness against different types of attacks, Tramer et al., \citep{tramer2019adversarial} extends adversarial training against multiple perturbations (ATMP) (typically the $\ell_1$, $\ell_2$, and $\ell_\infty$ attacks). Specifically, they consider two types of objective functions. The first one is the average of all perturbations (AVG), where the inner maximization problem of adversarial training finds adversarial examples for all types of attacks. The second one is worst-case perturbation (WST), where the inner maximization problem finds the adversarial examples with the largest loss within the union of the $\ell_p$ norm balls. Following these settings, researchers have proposed different algorithms to solve these two problems. Representative works are multi-steepest descent (MSD) \citep{maini2020adversarial}, and stochastic adversarial training (SAT) \citep{madaan2020learning}, which use different strategies to find adversarial examples in the $\ell_p$-norm balls and obtain some improvements comparing to the MAX and AVG.

However, there exist several crucial issues that are unsolved \emph{w.r.t.} ATMP. Firstly, the optimization process of ATMP is highly unstable compared to that of AT or standard training. Fig. \ref{fig-square} (c) and (d) give an example. The test robust accuracy fluctuates between different training epochs. Secondly, it is quite difficult to achieve a satisfying trade-off between different attacks. None of them achieves the best performance against all of the three attacks, as shown in Fig. \ref{fig-square} (b). Different algorithms tend to find sub-optimal around different local minima, resulting in a model perform well in one perturbation while worse defend against others. Last but most important, there is no theoretical study of ATMP currently. The exploration of ATMP methods is usually experimentally designed, without any theoretical guidelines.

In this work, we first study the smoothness and the loss landscape of ATMP. We show that the smoothness of $\ell_1$, $\ell_2$, and $\ell_\infty$ adversaries give different contributions to the smoothness of ATMP. It motivates us to study a question:
\begin{center}
    \emph{How to use the smoothness properties of different $\ell_p$ adversaries to design algorithms for ATMP?}
\end{center}

We study this question using the notion of uniform stability. In uniform stability analysis, excess risk, which is the sum of optimization error and generalization error, is highly related to the smoothness of the loss function. The formal and stability analysis and excess risk upper bound of ATMP are provided in Thm. \ref{thm:stdbound}. Inspired by the analysis, we propose adaptive smoothness-weighted adversarial training for multiple perturbations to improve the excess risk bound. Theoretically, our algorithms yields better bound (See our main results in Thm. \ref{gen} and Thm. \ref{opt}). Experimental results on CIFAR-10 and CIFAR-100 show that this technique mitigates the above issues and improves the performance of ATMP. Our solution achieves the state-of-the-art performance against the mixture of multiple perturbations attacks.

Our contributions are listed as follow: 
\begin{enumerate}
    \item We provide a comprehensive smoothness analysis of adversarial training for single and multiple perturbations. 
    \item We provide a uniform stability analysis on ATMP. Based on the analysis, we propose our stability-inspired algorithm: adaptive smoothness-weighted adversarial training for multiple perturbations.
    \item  Theoretically, our algorithms yields better excess risk bound. Experimentally, we obtain an improvement on robust accuracy, achieving the state-of-the-art performance on CIFAR-10 and CIFAR-100.
\end{enumerate}

\section{Related Work}
\label{rel}

In this section, we first introduce the standard adversarial training with a single type of perturbation, as well as its theoretical analysis.
We then introduce the adversarial training against multiple perturbations.
%
\paragraph{Adversarial training} 
Adversarial training (AT) has been demonstrated to be one of the most effective ways to increase the adversarial robustness \citep{szegedy2013intriguing}.
The key idea of AT is to augment the training set with adversarial examples during training.
Currently, most AT-based methods are trained with a single type of adversarial examples, and the $\ell_p$ (p=1, 2, or $\infty$) is commonly used to generate adversarial examples during training \citep{madry2017towards}.
It is shown that AT overfits the adversarial examples on the training set and generalizes badly on the test sets.
Many approaches have been proposed to increase the adversarial generalization \citep{raghunathan2019adversarial,schmidt2018adversarially}.
Meanwhile, there have been some attempts for the theoretical understanding of adversarial training, mainly focusing on the convergence properties and generalization bound.
For example, the work of \citep{gao2019convergence} studies the convergence of adversarial training in the neural tangent kernel (NTK) regime. 
Liu et al. study the smoothness of the loss function of adversarial training \citep{liu2020loss}. In terms of generalization bound, the work of \citep{yin2019rademacher,awasthi2020adversarial} study the generalization bound in terms of Rademacher complexity. 
The work of \citep{gao2019convergence} considers the VC-dimension bound of adversarial training. Xing et al. \citep{xing2021generalization} study the generalization of adversarial linear regression.

\paragraph{Adversarial robustness against multiple perturbations models} 
Recently, some works have demonstrated that adversarial training with a single type of perturbation cannot provide well defense against other types of adversarial attacks \citep{tramer2019adversarial} and several ATMP algorithms have been proposed accordingly \citep{maini2020adversarial,madaan2020learning,zhang2021composite,stutz2020confidence}.
The work of \citep{tramer2019adversarial} proposed to augment different types of adversarial examples into adversarial training and developed two augmentation strategies, \emph{i.e.}, MAX and AVG.
The MAX adopts the worst-case adversarial example among different attacks, while the AVG takes all types of adversarial examples into training.
Following the above pipeline, some later works developed different aggregation strategies (e.g., the MSD \citep{maini2020adversarial}, and SAT \citep{madaan2020learning}) for better robustness or training efficiency.
While these works can boost the adversarial robustness against multiple perturbations to some extent, the training process of ATMP is highly unstable, and there is no theoretical analysis about this.
The theoretical understanding of the training difficulty of ATMP is important for the further development of adversarial robustness for multiple perturbations.
Besides, there have also been some other works for adversarial robustness against multiple perturbations, such as {\it Ensemble models} \citep{maini2021perturbation,cheng2021mixture}, {\it Prepossessing} \citep{nandy2020approximate} and {\it Neural architectures search (NAS)} \citep{liu2020towards}.
The weakness of ensemble models or prepossess methods is that the performance is highly related to the classification quality or detection of different types of adversarial examples.
These methods either have lower performance or consider different tasks from the work we considered. Therefore, we mainly compare the algorithms MAX, AVG, MSD, and SAT in this work. 

\section{Preliminaries of Adversarial Training for Multiple Perturbations} 

Adversarial training is an approach to train a classifier that minimizes the worst-case loss within a norm-bounded constraint. Let $g(\theta,z)$ be the loss function of the standard counterpart. Given training dataset $\cS=\{z_i\}_{i=1\cdots n}$, the optimization problem of adversarial training is
\begin{equation}
\label{eq:at}
\min_\theta\frac{1}{n}\sum_{i=1}^{n}\max_{\|z_i-z_i'\|_p\leq\epsilon_p} g(\theta,z_i'),
\end{equation}

where $\epsilon_p$ is the perturbation threshold, $p=1,2$ or $\infty$ for different types of attacks. Usually, $g$ can also be written in the form of $\ell(f_{\theta}(\bx),y)$, where $f_\theta$ is the neural network to be trained and $(\bx,y)$ is the input-label pair. Adversarial training aims to train a model against a single type of $\ell_p$ attack. As AT with a single type of attacks may not be effecting under other types of attacks, adversarial training for multiple perturbations are proposed \citep{tramer2019adversarial}.
 Following the aforementioned literature, we consider the case that $p=1,2,\infty$. Two formulations can be use to tackle this problem.
\paragraph{Worst-case perturbation (WST)} The optimization problem of WST is formulated as follow,
\begin{equation}
\label{eq:max}
\begin{aligned}
\min_\theta\frac{1}{n}\sum_{i=1}^{n}\max_{p\in\{1,2,\infty\}}\max_{\|z_i-z_i'\|_p\leq\epsilon_p} g(\theta,z_i').
\end{aligned}
\end{equation}
WST aims to find the worst adversarial examples within the union of the three norm constraints for the inner maximization problem. The outer minimization problem updates model parameters $\theta$ to fit these adversarial examples. The MAX strategy \citep{tramer2019adversarial} are proposed for the optimization problem in Eq. (\ref{eq:max}). In each inner iteration, MAX takes the maximum loss on these three adversarial examples. Another algorithm for the optimization problem in Eq. (\ref{eq:max}) is multi-steepest descent (MSD)  \citep{maini2020adversarial}. In each PGD step in the inner iteration, MSD selects the worst among $\ell_1$, $\ell_2$, and $\ell_\infty$ attacks.

\paragraph{Average of all perturbations (AVG)} The optimization problem of AVG is formulated as follow
\begin{equation}
\label{eq:avg}
\begin{aligned}
\min_\theta\frac{1}{n}\sum_{i=1}^{n}\mathbb{E}_{p\sim\{1,2,\infty\}}\max_{\|z_i-z_i'\|_p\leq\epsilon_p} g(\theta,z_i'),
\end{aligned}
\end{equation}
where $p\sim\{1,2,\infty\}$ uniformly at random. The goal of the minimax problem in Eq. (\ref{eq:avg}) is to train the neural networks using data augmented with all three types of adversarial examples. The AVG strategy \citep{tramer2019adversarial} and the stochastic adversarial training (SAT) \citep{madaan2020learning} are two algorithms to solve the problem in Eq. (\ref{eq:avg}). In each inner iteration, AVG takes the average loss on these three adversarial examples and SAT randomly chooses one type of adversarial example among $\ell_1$, $\ell_2$, and $\ell_\infty$ attacks. 

Problem WST and AVG are similar but slightly different problems. WST aims to defend union attacks, \emph{i.e.,} the optimal attack within the union of multiple perturbations. AVG aims to defend mixture attacks, \emph{i.e.,} the attacker randomly pick one $\ell_p$ attack. In this paper, we mainly focus on problem AVG, we also discuss some solutions for problem WST.

\section{Smoothness Analysis}
We first study the smoothness of the minimax problems in Eq. (\ref{eq:max}) and (\ref{eq:avg}). To simplify the notation, let 
\begin{eqnarray*}
h^{p}(\theta,z)&=&\max_{\|z-z'\|_p\leq\epsilon_p}g(\theta,z'),\\ h^{avg}(\theta,z)&=&\mathbb{E}_{p\sim\{1,2,\infty\}}\max_{\|z-z'\|_p\leq\epsilon_p} g(\theta,z'),\\
h^{wst}(\theta,z)&=&\max_{p\in\{1,2,\infty\}}\max_{\|z-z'\|_p\leq\epsilon_p} g(\theta,z')
\end{eqnarray*}
be the loss function of standard adversarial training, worst-case multiple perturbation adversarial training, and average of all perturbations adversarial training, respectively. The population and empirical risks are the expectation and average of $h^{st}(\cdot)$, respectively. We use  $R_{\mathcal{D}}^{st}(\theta)$ and $R_{S}^{st}(\theta)$ to denote the population and empirical risk for adversarial training with different strategy, \emph{i.e.} $\text{st}\in\{1,2,\infty,\text{wst},\text{avg}\}$. 

\paragraph{Case study: Linear regression}
We use a simple case, adversarial linear regression, to illustrate the smoothness of the optimization problem of (\ref{eq:max}) and (\ref{eq:avg}). Let $f_\theta(\bx)= \theta^T\bx$ and $\ell(\theta^Tx,y)=|\theta^T\bx-y|^2$, we have the following proposition.
\begin{proposition}
\label{Prop1}
Let $\bX=[\bx_1,\cdots,\bx_n]^T$, $y=[y_1,\cdots,y_n]^T$, and $\delta=[\delta_1,\cdots,\\ \delta_n]^T$, we have
\begin{eqnarray*}
R_{S}^p(\theta)&=&[\|\bX\theta-y\|_2+\sqrt{n}\epsilon_p\|\theta\|_{p^*}]^2,\\
R_{S}^{wst}(\theta)&=&\max_{p\in\{1,2,\infty\}}[\|\bX\theta-y\|_2+\sqrt{n}\epsilon_p\|\theta\|_{p^*}]^2,\\
R_{S}^{avg}(\theta)&=&\mathbb{E}_{p\in\{1,2,\infty\}}[\|\bX\theta-y\|_2+\sqrt{n}\epsilon_p\|\theta\|_{p^*}]^2.
\end{eqnarray*}
\end{proposition}
The proof is deferred to \ref{A}. From Proposition \ref{Prop1}, the loss landscape of adversarial training is non-smooth because of the term $\|\theta\|_{p^*}$. Specifically, the loss function of $\ell_2$ adversarial training is non-smooth at $\theta= 0$. For $\ell_1$ adversarial training, the loss function is non-smooth at $|\theta_i|=|\theta_j|, \forall i,j$. For $\ell_\infty$ adversarial training, the loss function is non-smooth at $\theta_i=0, \forall i$. For adversarial training for multiple perturbations, the loss function is non-smooth at both $\theta_i=0, \forall i$ and $|\theta_i|=|\theta_j|, \forall i,j$. The non-smooth region of the loss function of ATMP is the union of that of the single perturbation cases. Different $\ell_p$ adversaries give different contribution to the smoothness of the loss function of ATMP.

\begin{figure}[htbp]
    \centering
	\includegraphics[width=4.5in]{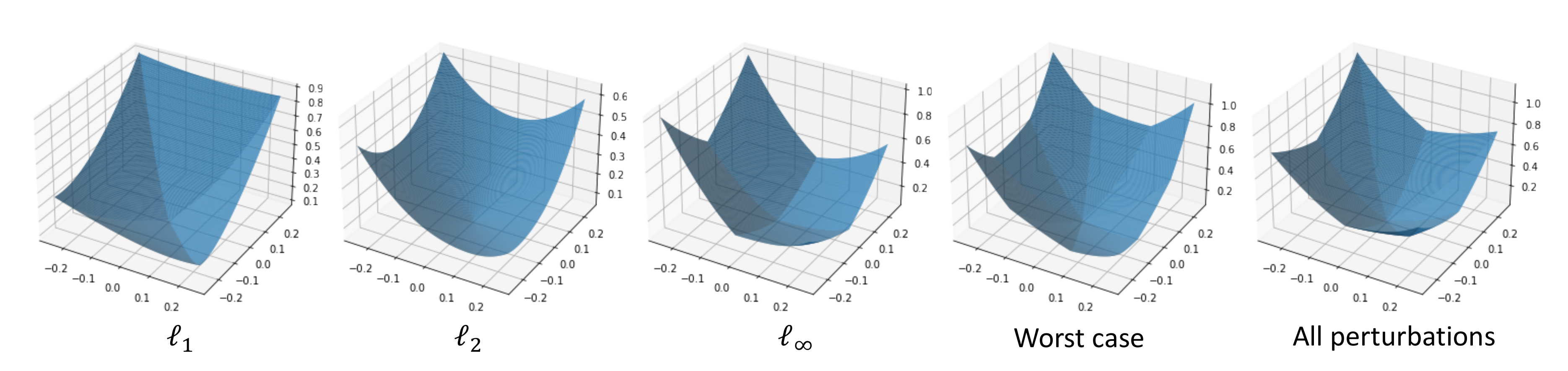}
	\caption{Loss landscape of adversarial linear regression for single and multiple perturbations.}
	\label{f1}
\end{figure}

In Fig. \ref{f1}, we give a numerical simulation and demonstrate the loss landscape in a two-dimensional case. In $\ell_2$ adversarial training, the loss landscape is smooth almost everywhere, except the original point. In $\ell_1$ and $\ell_\infty$ cases, the non-smooth region is a `cross'. In the cases of WST and AVG, the non-smooth region is the union of two `crosses'. 

\paragraph{General nonlinear model} Now let us consider general nonlinear models. Following the work of \citep{sinha2017certifiable}, without loss of generality, let us assume
\begin{assumption}
	\label{ass1}
	The function $g$ satisfies the following Lipschitzian smoothness conditions:
	\begin{equation*}
	\begin{aligned}
	&\|g(\theta_1,z)-g(\theta_2,z)\|\leq L\|\theta_1-\theta_2\|,\\
	&\|\nabla_\theta g(\theta_1,z)-\nabla_\theta g(\theta_2,z)\|\leq L_{\theta}\|\theta_1-\theta_2\|,\\
	&\|\nabla_\theta g(\theta,z_1)-\nabla_\theta g(\theta,z_2)\|\leq L_{\theta z}\|z_1-z_2\|,\\
	&\|\nabla_z g(\theta_1,z)-\nabla_\theta g(\theta_2,z)\|\leq L_{z\theta}\|\theta_1-\theta_2\|.\\
	\end{aligned}
	\end{equation*}
\end{assumption}
Assumption \ref{ass1} assumes that the loss function is smooth (in zeroth-order and first-order). While ReLU activation function is non-smooth, recent works \citep{allen2019convergence,du2019gradient} showed that the loss function of overparamterized DNNs is semi-smooth. It helps justify Assumption \ref{ass1}. Under Assumption \ref{ass1}, the following Lemma Provide the smoothness of ATMP.
\begin{lemmalist}
\label{lem:nonsmooth}
	Under Assumption \ref{ass1}, assuming in addtion that $g(\theta,z)$ is locally $\mu_p$-strongly concave for all $z\in\mathcal{Z}$ in $\ell_p$-norm. $\forall \theta_1, \theta_2$ and $\forall z\in\mathcal{Z}$, the following properties hold.
\begin{senenum}
\item \label{lem:1}
(Lipschitz function.) $\|h^{st}(\theta_1,z)-h^{st}(\theta_2,z)\|\leq L\|\theta_1-\theta_2\|$.
\item \label{lem:3} (Gradient Lipschitz.) If we  Then, $\|\nabla_\theta h^{st}(\theta_1,z)-\nabla_\theta h^{st}(\theta_2,z)\|\leq \beta_{st}\|\theta_1-\theta_2\|$, where
\begin{equation*}
    \beta_{st}=\begin{cases}
    L_{\theta z}L_{z\theta}/\mu_{st}+L_\theta & \text{st}\in\{1,2,\infty\},\\
    L_{\theta z}L_{z\theta}/\min_p\mu_p+L_\theta & \text{st}=\text{wst},\\
    \mathbb{E}_{p\sim\{1,2,\infty\}}\beta_p & \text{st}=\text{avg}.
    \end{cases}
\end{equation*}
\end{senenum}
\end{lemmalist}

Proof: see \ref{a2}. Lemma \ref{lem:nonsmooth} shows that adversarial surrogate loss in different $\ell_p$ adversaries have different smoothness in strongly-concave case. This property is not specific for strongly-concave case. The analysis of non-strongly-concavity case is provided in \ref{a3}. Lemma \ref{lem:nonsmooth} motivates us to study a question:
\begin{center}
    \emph{How to achieve better performance on adversarial robustness against different $\ell_p$ adversarial attacks utilizing the smoothness properties?}
\end{center}
In the next section, we first discuss the stability analysis of ATMP. Then, we discuss how to utilize the smoothness-properties of different $\ell_p$ adversaries to obtain smaller generalization bound.

\section{Stability-based Excess Risk Analysis}
In this section, we focus on the problem AVG. Assuming that the target model are facing $P$ potential attacks. In the above setting, $P=3$ and the three attacks are $\ell_1$, $\ell_2$, and $\ell_\infty$ attacks. The test and training performance against mixture attacks is

$$ R_{\cal D}(\theta)=\frac{1}{P}\sum_{p=1}^P R_{\cal D}^p(\theta)\ \  \text{and}\ \ R_{S}(\theta)=\frac{1}{P}\sum_{p=1}^P R_{S}^p(\theta),$$
respectively.

\paragraph{Risk Decomposition} Let $\theta^*$ and $\bar\theta$ be the optimal solution of $R_{\cal D}(\theta)$ and $R_S(\theta)$, respectively. Then for the algorithm output $\hat\theta=A(S)$, the excess risk can be decomposed as
\begin{equation}
\begin{aligned}
    &R_{\cal D}(\hat\theta)-R_{\cal D}(\theta^*)\\
    =&\underbrace{R_{\cal D}(\hat\theta)-R_{S}(\hat\theta)}_{\mathcal{E}_{gen}}
    +\underbrace{R_{S}(\hat\theta)-R_{S}(\bar\theta)}_{\mathcal{E}_{opt}}
    +&\underbrace{R_{S}(\bar\theta)-R_{S}(\theta^*)}_{\leq 0}+\underbrace{R_{S}(\theta^*)-R_{\cal D}(\theta^*)}_{\mathbb{E}=0}.
    \end{aligned}
\end{equation}
To control the excess risk, we need to control the generalization gap $\mathcal{E}_{gen}$ and the optimization gap $\mathcal{E}_{opt}$. In the rest of the paper, we use $\mathcal{E}_{gen}$ and $\mathcal{E}_{opt}$ to denote the \emph{expectation} of the generalization and optimization gap. The smoothness of the loss function is highly related to the generalization gap $\mathcal{E}_{gen}$ and the optimization gap $\mathcal{E}_{opt}$, we first provide the the optimization error bound \citep{nemirovski2009robust} and stability-based generalization bound\footnote{We refer the readers to \citep{hardt2016train} for the preliminaries of uniform stability.} \citep{hardt2016train} for running SGD on Eq. (\ref{eq:avg}).

\begin{theorem}
\label{thm:stdbound}
	Under Assumption \ref{ass1}, assuming in addition that $g(\theta,z)$ is convex in $\theta$ for all given $z\in\mathcal{Z}$. Let $D=\|\theta^0-\theta^*\|$, where $\theta^0$ is the initialization of SGD. Suppose that we run SGD with step sizes $\alpha\leq 1/\beta_{avg}$ for $T$ steps. Then, adversarial training satisfies
\begin{equation}
    \mathcal{E}_{opt}\leq \frac{D^2+L^2T\alpha^2}{2T\alpha},\ \ \mathcal{E}_{gen}\leq\frac{2L^2T\alpha}{n}.\footnote{For varing stepsize, we can replace $T\alpha$ and $T\alpha^2$ by $\sum_{t=1}^T\alpha_t$ and $\sum_{t=1}^T\alpha_t^2$, respectively.}
\end{equation}
\end{theorem}

Therefore, the $\beta$-gradient Lipschitz constant of the loss function is related to the choice of stepsize, the optimization and generalization bound. In the loss function of AVG, each of the $\ell_p$ adversarial loss have different Lipschitz constant $\beta_p$. It motivates us to study whether we can assign different stepsize to different $\ell_p$ adversarial loss to improve the excess risk.

\subsection{Smoothness-weighted Adversarial Training for Multiple Perturbations}
Considering the algorithm
\begin{equation}\label{alg1}
    \theta^{t+1}=\theta^t-\frac{1}{P}\bigg[\alpha_1^t\nabla R_{S}^1(\theta^t)+\cdots +\alpha_P
    ^t \nabla R_{S}^P(\theta^t)\bigg],
\end{equation}
In each of the iterations $t$, we assign stepsize $\alpha_p^t$ to the $p^{th}$-tasks.
\paragraph{Properties of Update Rules} We define $G_{z}(\theta)=\theta-\frac{1}{P}\sum_{p=1}^P\alpha_p\nabla h^p(\theta,z)$ be the update rule. The following lemma holds.

\begin{lemma}[Non-expansive]
\label{updaterules}
	Assuming that the function $h^p(\theta,z)$ is $\beta_p$-gradient Lipschitz, convex for all $z\in\mathcal{Z}$. Then, $\forall \theta_1, \theta_2$ and $\forall z\in\mathcal{Z}$, for $\alpha_p\leq 1/\beta_p$, we have $\|G_z(\theta_1)-G_z(\theta_2)\|\leq \|\theta_1-\theta_2\|.$
\end{lemma}
Proof of Lemma \ref{updaterules} is deferred to \ref{proofl1}. Based on Lemma \ref{updaterules}, we have the following generalization guarantee for problem AVG.
\begin{theorem}[Generalization bounds of smoothness-weighted learning rate] 
\label{gen}
	Under Assumption \ref{ass1}, assume in addition that $h^p(\theta,z)$ is convex in $\theta$ for all given $z\in\mathcal{Z}$. Suppose that we run Algorithm \ref{alg1} with step sizes $\alpha_p\leq 1/\beta_p$ for $T$ steps. Then, adversarial training satisfies
uniform stability with
\[
\mathcal{E}_{gen}\leq\frac{2L^2T\sum_{p=1}^P\alpha_p/P}{n}.
\]
\end{theorem}
Proof: The proof is based on Lemma \ref{updaterules} and defer to \ref{a4}.

Let $\alpha_{sw}^t=(\alpha_1^t+\cdots+\alpha_P^t)/P$, we have $\mathcal{E}_{gen}\leq 2L^2T\alpha_{sw}/n$.
\begin{theorem}[Optimization bounds of smoothness-weighted learning rate] 
\label{opt}
	Under Assumption \ref{ass1}, assume in addition that $h^p(\theta,z)$ is convex in $\theta$ for all given $z\in\mathcal{Z}$. Suppose that we run SGD with step sizes $\alpha_p^t\leq 1/\beta_p$ for $T$ steps. Then, adversarial training satisfies
\[
\mathcal{E}_{opt}\leq \frac{D^2+L^2T\alpha_{sw}^2}{2T\alpha_{sw}}+B\frac{\sum_{p=1}^P|\alpha_{sw}-\alpha_p|}{\alpha_{sw}}.
\]
\end{theorem}
The proof is deferred to \ref{a5}. The first term has the same form as Theorem \ref{thm:stdbound}, the second term is an additional bias term introduced by the smoothness-weighted learning rate. Combining the $\mathcal{E}_{opt}$ and $\mathcal{E}_{gen}$, we have
\[
\mathcal{E}_{gen}+\mathcal{E}_{opt}\leq \underbrace{\frac{2L^2T\alpha_{sw}}{n}+\frac{D^2+L^2T\alpha_{sw}^2}{2T\alpha_{sw}}}_{\text{The same as Thm. \ref{thm:stdbound} with different $\alpha$}}+\underbrace{B\frac{\sum_{p=1}^P|\alpha_{sw}-\alpha_p|}{\alpha_{sw}}}_{\text{bias term}}.
\]

Optimizing the first two terms with respective to $\alpha$, we have
\[
\alpha^*=\frac{D\sqrt{n}}{L\sqrt{T(n+2T)}}.
\]
In adversarial training, $T$ cannot be too large because of robust overfitting. Then, The right-hand-side may be too large and we may not choose $\alpha^*$ as the learning rate. We need a larger $\alpha$ to reduce the first two term. From the previous dicussion, we have
\begin{equation*}
    \alpha_{avg}\leq \frac{P}{\beta_1+\cdots+\beta_P}\ and \ \alpha_{sw}\leq \frac{\frac{1}{\beta_1}
    +\cdots+\frac{1}{\beta_P}}{P}.
\end{equation*}

Therefore, $\alpha_{sw}$ can be view as the inverse of the harmonic mean of $\beta_p$ and $\alpha_{avg}$ can be view as the inverse of the arithmetic mean of $\beta_p$. $\alpha_{sw}$ is larger and reduce the first two terms when $T$ is small.

Overall, we need smaller $T$ and carefully chosen learning rates to speed up adversarial training to avoid robust overfitting. Smoothness-weighted learning rate gives us a way to increase the learning rate. As a side-effect, it introduce an additional bias term. In experiments, we will show that $\alpha_{sw}$ can improve the test performance.

\subsection{Proposed Method}
In practice, $\beta_p$ is unknown. We adaptively estimate $\beta_p$ in each iterations. By descent Lemma, we have
\[
h^p(\theta_t)-h^p(\theta^*)\leq \nabla h^p(\theta^*)\langle\theta_t-\theta^*\rangle+ \frac{\beta_p}{2}\|\theta_t-\theta^*\|^2= \frac{\beta_p}{2}\|\theta_t-\theta^*\|^2.
\]
It suggests us to use the following rules 

\[
\alpha_p^t\propto\frac{1}{\beta_p}\propto\frac{\|\theta_t-\theta^*\|^2}{h^p(\theta_t)-h^p(\theta^*)}\propto \frac{1}{h^p(\theta_t)-h^p(\theta^*)}
\]
to approximate $\beta_p$ and to choice $\alpha_p^t$. Assuming that $h^p(\theta^*)=0$, we can use $\sum_{p}h^p(\theta_t)/(P h^p(\theta_t))$ as the weight of $\alpha_p^t$. Given the initial learning rate schedule $\alpha^1,\cdot,\alpha^T$, the following Algorithm \ref{alg1} is the proposed adaptive smoothness-weighted ATMP.
\begin{algorithm}
\caption{Adaptive Smoothness-Weighted ATMP}
\label{alg1}
\begin{algorithmic}
    \STATE \textbf{Inputs:} classifier $f_\theta(\bx,y)$, dataset $\{\bx_i,y_i\}_{i=1,\cdots,n}$.
    \STATE \textbf{Initialize} learning rate schedule $\alpha^1,\cdot,\alpha^T$. \\
    \FOR{$t=1$ to $T$}
    \FOR{$p=1$ to $P$}
    \STATE\begin{equation*}
\begin{aligned}
\mathcal{L}_p=\frac{1}{n}\sum_{i=1}^{n}\max_{\|\delta_i\|_p\leq\epsilon_p}\ell(f_{\theta_t}(\bx_i+\delta_i,y_i).
\end{aligned}
\end{equation*}
    \ENDFOR
    \STATE Define $\mathcal{L}= \mathcal{L}_1+\cdots+ \mathcal{L}_P$.
    \STATE Update $\alpha_p^t=\alpha^t\times\mathcal{L}/P\mathcal{L}_P$, $p=1,2,\cdots,P$.
    \STATE Update 
    \begin{equation*}
    \theta^{t+1}=\theta^t-\frac{1}{P}\bigg[\alpha_1^t\nabla R_{S}^1(\theta^t)+\cdots +\alpha_P
    ^t \nabla R_{S}^P(\theta^t)\bigg],
\end{equation*}

    \ENDFOR
\end{algorithmic}
\end{algorithm}

\section{Experiments}

\begin{table*}[htbp]
    \centering
    \caption{Test accuracy (\%) of different algorithms (MAX, AVG, MSD, SAT, and ADT) against $\ell_1$, $\ell_2$, and $\ell_\infty$ attacks on CIFAR-10.}
    \resizebox{\linewidth}{!}{%
    \begin{tabular}{c|c|cccccc}
    \hline
      \multicolumn{2}{c|}{Dataset}&\multicolumn{6}{c}{CIFAR-10}\\ \hline
         \multicolumn{2}{c|}{Attack methods}& clean & $\ell_1$ & $\ell_2$ & $\ell_\infty$& Union & Mix \\ \hline
        \multirow{3}{*}{AT}&$\ell_1$ & 93.22 & 89.81 & 0.00 & 0.00 &0.00&29.98\\ 
        &$\ell_2$ & 88.66 & 41.41 & 61.72 & 39.84 &18.41&47.66\\
        &$\ell_\infty$ & 84.94 & 17.19 & 53.91 & 46.88 &40.11&39.32 \\ \hline
        \multirow{2}{*}{WST (Eq. \ref{eq:max})}&MAX & 84.96 & 52.63 & 64.74 & 46.93&46.08$\pm$0.43& 54.77$\pm$0.22\\ 
        &MSD & 83.51 & 54.92 & 67.68 & 49.88 & 46.99$\pm$0.23&57.49$\pm$0.11\\ \hline
        \multirow{3}{*}{AVG (Eq. \ref{eq:avg})}&AVG & 85.28 & 58.78 & 68.08 & 43.87&43.28$\pm$0.55&56.91$\pm$0.34\\
        &SAT & 85.23 & 58.68 & 67.77 & 43.59&43.12$\pm$1.89&56.68$\pm$1.01\\
        &ADT&85.87&\textbf{61.81}&\textbf{69.61}&\textbf{46.64}&\textbf{46.05}$\pm$0.31&\textbf{59.16}$\pm$0.09\\
        \hline
    \end{tabular}
    \label{table1}}
\end{table*}

\subsection{Performance of ADT}
\paragraph{Datasets and Classification Models} We conduct experiments on two widely used benchmark datasets: CIFAR-10 and CIFAR-100 \citep{krizhevsky2009learning}. CIFAR-10 includes 50k
training images and 10k test images with 10 classes.  CIFAR-100 includes 50k training images and 10k test images with 100 classes. For classification models, we use the pre-activation version of the ResNet18 architecture (Pre-Res18) \citep{he2016identity}.  

\paragraph{Evaluation Protocol} We consider two formulations, WST in Eq. (\ref{eq:max}) and AVG in Eq. (\ref{eq:avg}). We use ADT to stand for our proposed algorithm. Since ADT is designed for defending mixture attacks, we mainly use Mix to evaluate the performance. We also provide the performance against union attacks.

\paragraph{Training settings} We adopt popular training techniques and three widely considered types of adversarial examples mentioned above: $\ell_1$, $\ell_2$, and $\ell_{\infty}$ attacks in the inner maximization. For $\ell_1$ attack,  we adopt the attack method used in \citep{maini2020adversarial}. For $\ell_2$ and $\ell_{\infty}$, we utilize the multi-step PGD attack methods \citep{madry2017towards}. The perturbation budgets are set as 12, 0.5, 0.03. For better convergence performance of the inner maximization problem \citep{tramer2020adaptive}, we set the number of steps as 50 and further increase it to 100 in the testing phase. For the stepsize in the inner maximization, we set it as 1, 0.05, and 0.003. respectively. \emph{Cyclic Learning Rates:} in the outer minimization, we use the SGD optimizer with momentum 0.9 and weight decay $5 \times 10^{-4}$, along with a variation of learning rate schedule from \citep{smith2018disciplined}, which is piece-wise linear from 0 to 0.1 over the first 40 epochs, down to 0.005 over the next 40 epochs, and finally back down to 0 in the last 20 epochs. \emph{Stochastic Weight Averaging and Early Stopping:} following the state-of-the-art training techniques for adversarial training, we incorporate stochastic weight averaging (SWA) \citep{izmailov2018averaging} and early stopping in ATMP. It is shown that SWA could find flat local minima and yields performance \citep{stutz2021relating}. The update of SWA is $\theta_{swa}^t=\gamma  \theta_{swa}^{t-1}+(1-\gamma)\theta^{t-1}$,
where $\gamma$ is a hyper-parameter and the final $\theta_{swa}^T$ is used for evaluation.
We follow the setting of \citep{izmailov2018averaging} and start SWA from the 60-th epoch for all the methods we compare. For all the methods, we repeat five runs.

\paragraph{Comparison of AVG, SAT, and ADT} We mainly compare the three methods for the problem in Eq. (\ref{eq:avg}). The highest numbers are in bold. Mixture attack (average of all attacks) is the main index to evaluate the performance. On CIFAR-10, we can see that ADT achieve the highest robust accuracy in 59.16\%. On CIFAR-100, ADT achieves 35.39\% robust accuracy. The performance proves that ADT can automatically be adaptive to the ones that have not been optimized well. Furthermore, ADT has a smaller deviation than SAT and AVG. 

\subsection{Discussion: different goals of WST and AVG}

\paragraph{Comparison of ADT and MSD} The formulations of Eq. (\ref{eq:avg}) and (\ref{eq:max}) have similar but slightly different goals. WST tries to fit the adversarial examples who have the largest loss within the union of the three norms. AVG is designed to defend the mixture attacks. The difference in ADT and MSD is the difference in the optimization problems AVG and WST. In Table \ref{table1}, ADT achieves comparable performance, 46\%, on union of all the attacks. In terms of mixture attack, ADT achieves 59\% robust accuracy, while the robust accuracy of MSD is 57\%.
\begin{table*}[htbp]
    \centering
    \caption{Test accuracy (\%) of different algorithms (MAX, AVG, MSD, SAT, and ADT) against $\ell_1$, $\ell_2$, and $\ell_\infty$ attacks on CIFAR-100.}
    \resizebox{\linewidth}{!}{%
    \begin{tabular}{c|c|cccccc}
    \hline
      \multicolumn{2}{c|}{Dataset}&\multicolumn{6}{|c}{CIFAR-100}\\ \hline
         \multicolumn{2}{c|}{Attack methods}& clean &  $\ell_1$  & $\ell_2$ & $\ell_\infty$ & Union &Mix  \\ \hline
        \multirow{3}{*}{AT}&$\ell_1$ &  70.98 & 73.44 & 00.82 & 00.51 &0.04 &24.92\\ 
        &$\ell_2$ & 63.76 & 21.88 & 43.75 & 20.31 &12.07&28.64\\
        &$\ell_\infty$ & 58.86 & 11.02 & 39.22 & 28.01 & 9.41&26.08\\ \hline
        \multirow{2}{*}{WST (Eq. \ref{eq:max})}&MAX & 57.82 & 30.36 & 40.71 & 26.08 &25.03$\pm$0.39& 32.38$\pm$0.18\\ 
        &MSD & 57.33 & 32.08 & 41.90 & 27.06 &26.21$\pm$0.22&34.02$\pm$0.08\\ \hline
        \multirow{3}{*}{AVG (Eq. \ref{eq:avg})}&AVG &  59.75 & 35.55 & 41.03 & 24.61 & 24.31$\pm$0.68 & 33.73$\pm$0.41\\
        &SAT &  59.25 & 35.60 & 42.33 & 25.01 & 24.78 $\pm$1.41&34.31$\pm$0.88\\
        &ADT&   59.41& \textbf{37.64}&\textbf{42.82}&\textbf{25.70}&\textbf{25.29}$\pm$0.15& \textbf{35.39}$\pm$0.07\\
        \hline
    \end{tabular}}
    \label{table1}
\end{table*}  
Overall, $\ell_\infty$ adversarial examples induce larger norm within the union of the three norms, and MSD tends to find and fit them. ADT pays more attention to $\ell_1$ adversarial examples. Comparing the overall performance, ADT achieves better robustness trade-off against mixture adversarial attacks.

\begin{figure*}[htbp]
\centering
\scalebox{0.9}{
\subfigure[]{
\begin{minipage}[htp]{0.24\linewidth}
\centering
\includegraphics[width=1.2in]{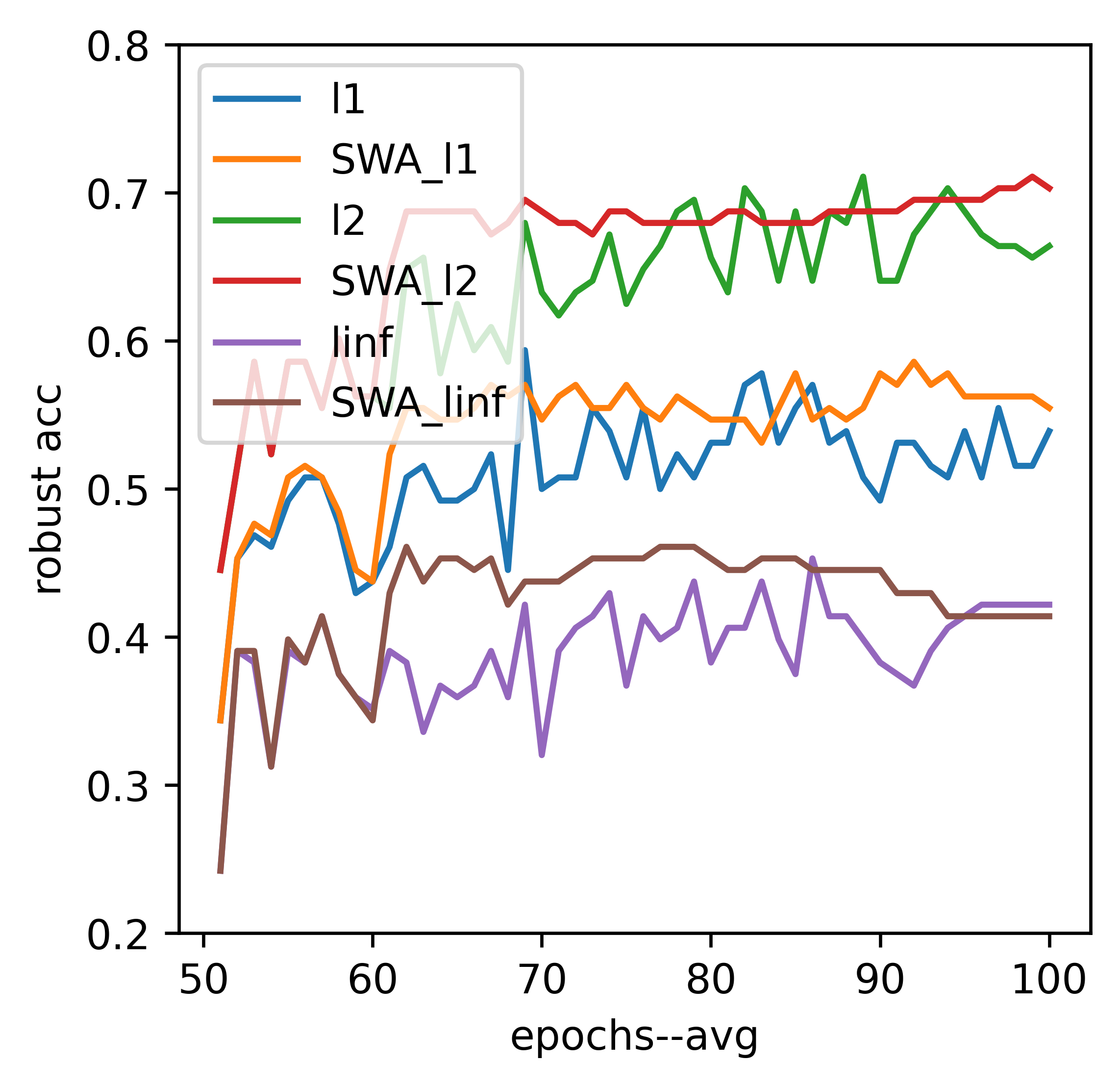}
\end{minipage}%
}
\subfigure[]{
\begin{minipage}[htp]{0.24\linewidth}
\centering
\includegraphics[width=1.2in]{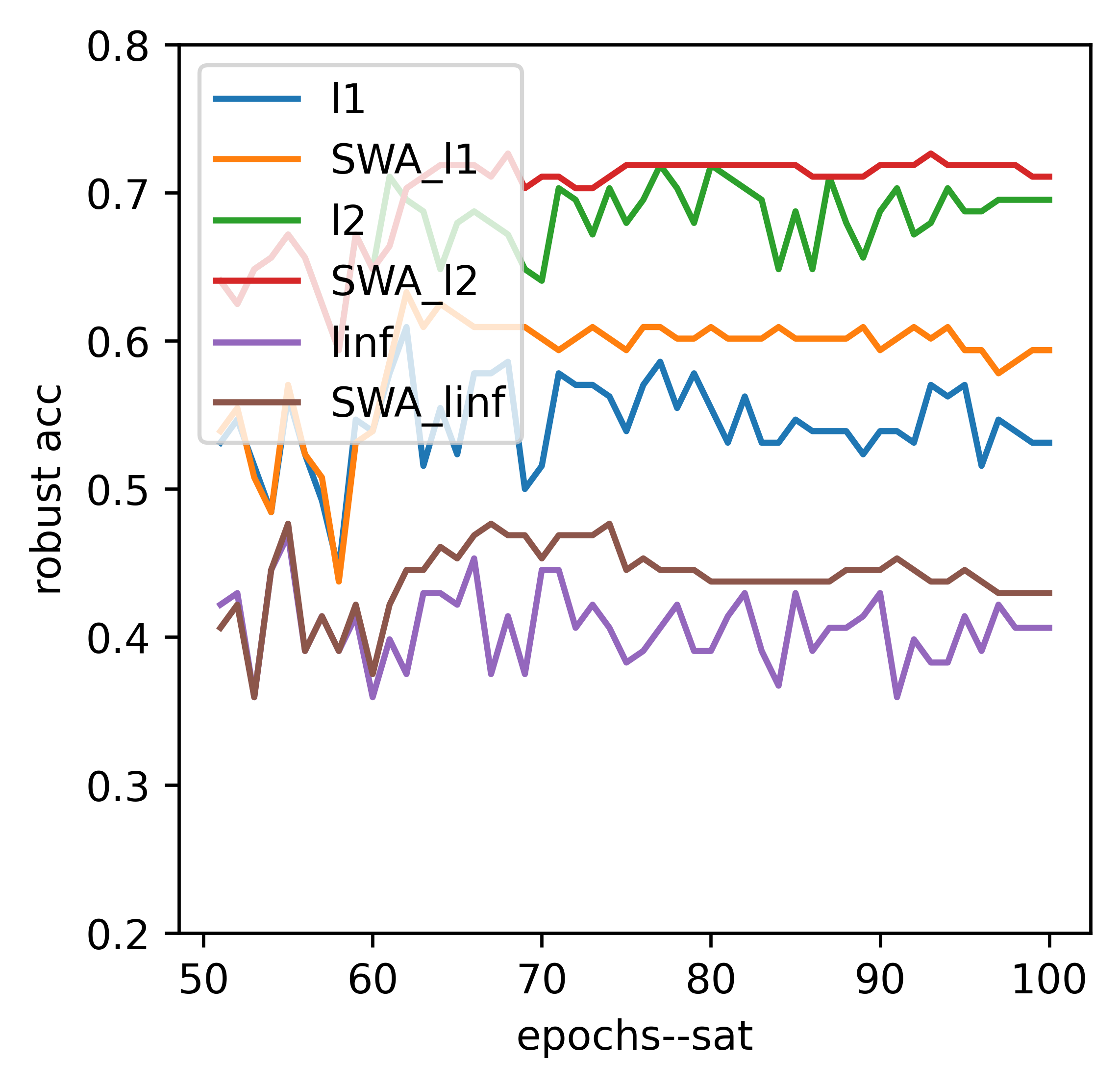}
\end{minipage}
}
\subfigure[]{
\begin{minipage}[htp]{0.24\linewidth}
\centering
\includegraphics[width=1.2in]{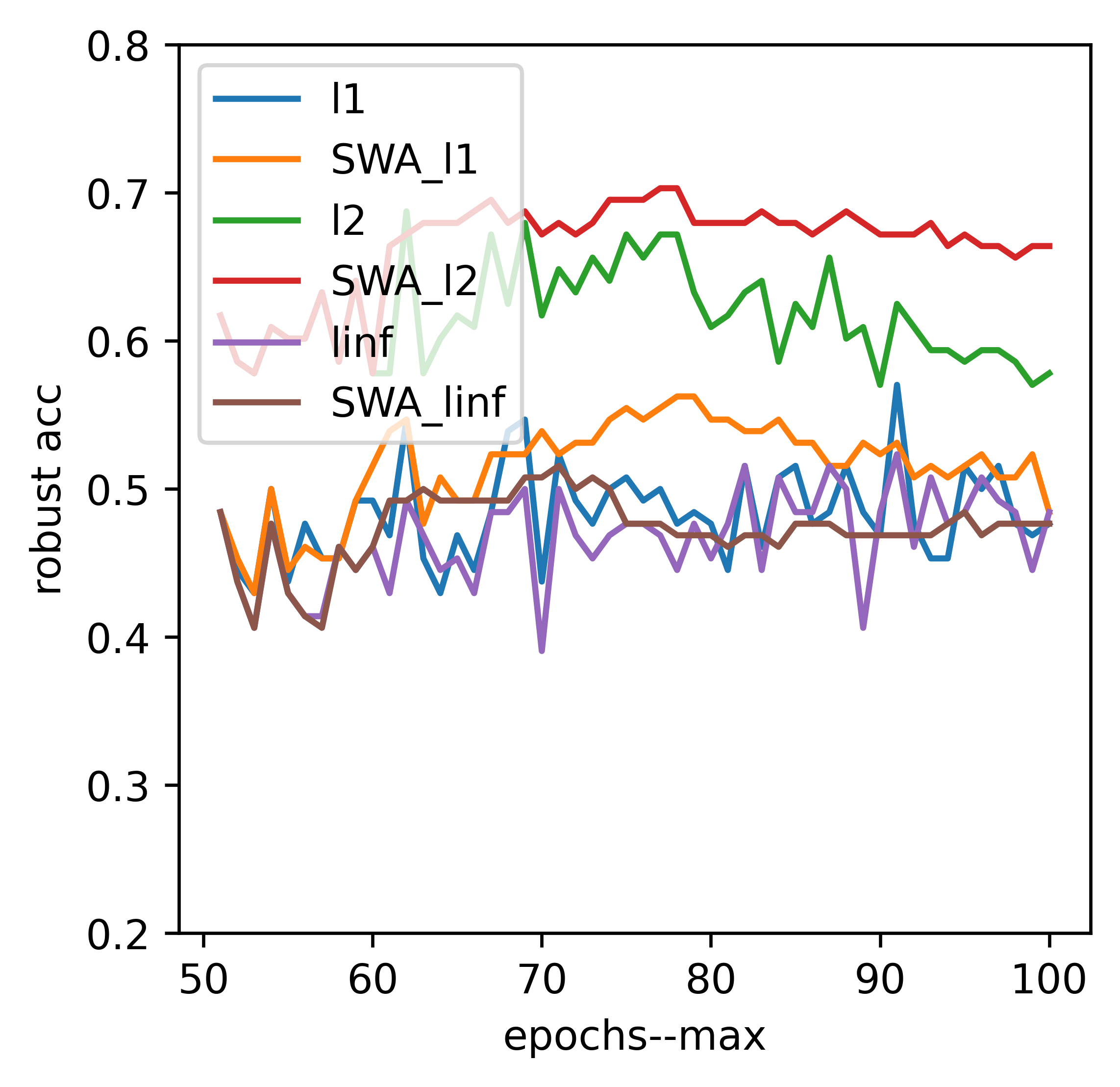}
\end{minipage}
}
\subfigure[]{
\begin{minipage}[htp]{0.24\linewidth}
\centering
\includegraphics[width=1.2in]{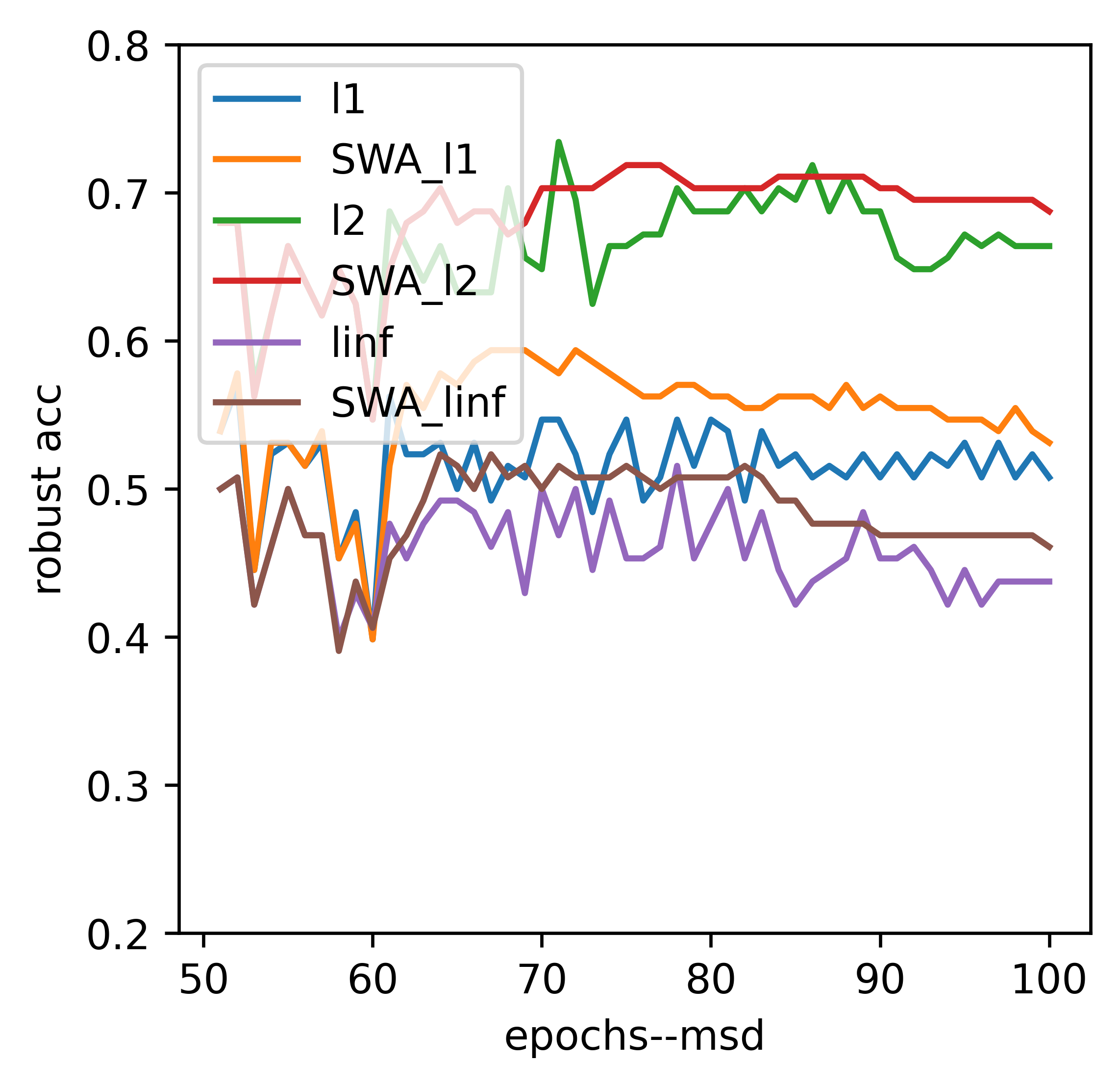}
\end{minipage}
}}
\centering
\caption{Tendency curves of robust accuracy against different types of adversarial attacks. The models are trained by ATMP using four different strategies, with and without SWA.}
\label{cifar10}
\end{figure*}

\paragraph{Solutions for WST} Our paper mainly focus on problem AVG. We also discuss some solutions to improve the performance of WST. In Fig. \ref{cifar10}, we plot the robust accuracy against $\ell_1$, $\ell_2$, and $\ell_\infty$ adversarial attacks of four different strategies (MAX, AVG, MSD and SAT) on CIFAR-10. It shows that SWA is an effective methods to improve the performance of ATMP. For instance, in subplot (a), the l1 / l2 / linf denotes the robust accuracy of ATMP trained with AVG strategy against the $\ell_1$ / $\ell_2$ / $\ell_{\inf}$ adversarial attacks.
While the SWA\_l1 / SWA\_l2 / SWA\_linf relates to the ATMP model that trained by AVG strategy coupled with SWA. From the plots, we observe that the test accuracy is highly unstable among different training epochs without SWA. When coupling with SWA, the tendency curves of all four ATMP strategies are largely stabilized. Using early stopping, we could find the checkpoint for the best performance. On CIFAR-10, the improvement of SWA is 1.82\%, 3.39\%, 1.56\%, and 2.67\% using MSD, SAT, AVG, and MAX, respectively. Other training techniques are discussed in \ref{b3}.

\section{Conclusion}
In this paper, we try to find out the difficulty of adversarial training for multiple perturbations from the perspective of optimization. Specifically, we study the smoothness of the loss function of ATMP and provide a stability analysis. Based on the analysis, we propose adaptive smoothness-weighted adversarial training for multiple perturbations, which achieve better generalization bound and achieve state-of-the-art performance against multiple perturbations.

\clearpage
\bibliography{main.bib}
\bibliographystyle{apalike}
\clearpage

\appendix
\section{Proof of Theorem}
\label{A}
In this section, we provide the detailed proof.
\subsection{Proof of Proposition \ref{Prop1}}
\label{A1}
Since 
\begin{eqnarray*}
   & &\|(\bX+\delta)\theta-y\|^2\\
    &\leq& [\|\bX\theta-y\|+\|\delta\theta\|]^2\\
    &= & [\|\bX\theta-y\|+\sqrt{\sum|\delta_i\theta|^2}]^2\\
    &\leq & [\|\bX\theta-y\|+\sqrt{\sum[\|\delta_i\|_p\|\theta\|_{p*}]^2}]^2\\
    &\leq & [\|\bX\theta-y\|+\sqrt{n\epsilon_p^2\|\theta\|_{p*}^2}]^2\\
    &=&[\|\bX\theta-y\|_2+\sqrt{n}\epsilon_p\|\theta\|_{p^*}]^2,
\end{eqnarray*}
where the first inequality is due to triangle inequality, the second inequality is due to Cauchy-Schwarz inequality, and the last inequality is due to the constraint $\|\delta\|_{p,\infty}\leq\epsilon_p$. Choosing $\delta_i$ to satisfy the aforementioned three inequalities, we obtain 
\begin{eqnarray*}
R_{S}^p(\theta)=\max_{\|\delta\|_{p,\infty}\leq\epsilon_p}\|(\bX+\delta)\theta-y\|^2
=[\|\bX\theta-y\|_2+\sqrt{n}\epsilon_p\|\theta\|_{p^*}]^2.
\end{eqnarray*}
Based on this, we directly have 
\begin{eqnarray*}
R_{S}^{wst}(\theta) &=&\max_{p\in\{1,2,\infty\}}\max_{
	\|\delta\|_{p,\infty}\leq\epsilon_p}\|(\bX+\delta)\theta-y\|^2\\
	&=&\max_{p\in\{1,2,\infty\}}[\|\bX\theta-y\|_2+\sqrt{n}\epsilon_p\|\theta\|_{p^*}]^2,
\end{eqnarray*}
\begin{eqnarray*}R_{S}^{avg}(\theta)&=&\mathbb{E}_{p\in\{1,2,\infty\}}\max_{\|\delta\|_{p,\infty}\leq\epsilon_p}\|(\bX+\delta)\theta-y\|^2\\&=&\mathbb{E}_{p\in\{1,2,\infty\}}[\|\bX\theta-y\|_2+\sqrt{n}\epsilon_p\|\theta\|_{p^*}]^2.\end{eqnarray*}\qed
\subsection{Proof of Lemma \ref{lem:nonsmooth}}\label{a2}
Proof:\\
Case 1: $\text{st}\in\{1,2,\infty\}$. The proof can be found in \citep{sinha2017certifiable,wang2019convergence}.\\
Case 2: $\text{st}=wst$. Since $g(\theta,z)$ is locally $\mu_p$-strongly concave for all $z\in\mathcal{Z}$ in $\ell_p$-norm, $g(\theta,z)$ is locally $\min\mu_p$-strongly concave within the union of $\ell_p$ norm ball for all $z\in\mathcal{Z}$.\\
Case 3:  $\text{st}=avg$. Since $h^{avg}(\theta,z)=\mathcal{E}h^p(\theta,z)$, we have
\[
\|\nabla_\theta h^{avg}(\theta_1,z)-\nabla_\theta h^{avg}(\theta_2,z)\|=\|\mathbb{E}(\nabla_\theta h^{p}(\theta_1,z)-\nabla_\theta h^{p}(\theta_2,z))\|\leq \mathbb{E} \beta_p \|\theta_1-\theta_2\|.
\]\qed
\subsection{Discussion on Non-Strongly-Convex Cases}
\label{a3}
\begin{assumption}
	\label{ass2}
	The function $g$ satisfies the following Lipschitzian smoothness conditions:
	\begin{equation*}
	\begin{aligned}
	&\|g(\theta_1,z)-g(\theta_2,z)\|\leq L\|\theta_1-\theta_2\|,\\
	&\|\nabla_\theta g(\theta_1,z)-\nabla_\theta g(\theta_2,z)\|\leq L_{\theta}\|\theta_1-\theta_2\|,\\
	&\|\nabla_\theta g(\theta,z_1)-\nabla_\theta g(\theta,z_2)\|\leq L_{\theta z}^p\|z_1-z_2\|_p,\\
	&\|\nabla_z g(\theta_1,z)-\nabla_\theta g(\theta_2,z)\|\leq L_{z\theta}\|\theta_1-\theta_2\|.\\
	\end{aligned}
	\end{equation*}
\end{assumption}
Assumption \ref{ass2} assumes that the gradient Lipschitz in different $\ell_p$-norm are $L_{\theta z}^p$, which can be verified by the relation between norms.
\begin{lemmalist}
\label{lem:nonsmooth2}
	Under Assumption \ref{ass2}, $\forall \theta_1, \theta_2$ and $\forall z\in\mathcal{Z}$, the following properties hold.
\begin{senenum}
\item \label{lem:1}
(Lipschitz function.) $\|h^{st}(\theta_1,z)-h^{st}(\theta_2,z)\|\leq L\|\theta_1-\theta_2\|$.

\item \label{lem:2} (Non-gradient Lipschitz.) $\|\nabla_\theta h^{st}(\theta_1,z)-\nabla_\theta h^{st}(\theta_2,z)\|\leq L_{\theta}\|\theta_1-\theta_2\|+\eta_{st}$, where $\eta_{p}=2L_{\theta z}^p\epsilon_p$, $\eta_{wst}=2\max\{L_{\theta z}^p\epsilon_p\}$, and $\eta
_{avg}=2\mathbb{E}_pL_{\theta z}^p\epsilon_p$.
\end{senenum}
\end{lemmalist}

Lemma \ref{lem:2} and \ref{lem:3} show that adversarial surrogate loss in different $\ell_p$ adversaries have different smoothness in general non-concave case.

Proof: Notice that $R_{S}^{st}(\theta)=\sum_{i=1}^n h^{st}(\theta,z_i)/n$, we only need to prove that $\forall \bx$, we have 
	\begin{equation}
	\label{gst}
\begin{aligned}
&\|h^{st}(\theta_1,z)-h^{st}(\theta_1,z)\|\leq L_{\theta}\|\theta_1-\theta_2\|,\\
&\|\nabla_\theta h^{st}(\theta_1,z)-\nabla_\theta h^{st}(\theta_2,z)\|\leq L_{\theta\theta}\|\theta_1-\theta_2\|+\eta_{st},
\end{aligned}
\end{equation}
where $\text{st}\in\{1,2,\infty,\text{wst},\text{avg}\}$ with $\eta_{p}=2L_{\theta\bx}^p\epsilon_p$, $\eta_{wst}=2\max\{L_{\theta\bx}^p\epsilon_p\}$, and $\eta
_{avg}=2\mathbb{E}_pL_{\theta\bx}^p\epsilon_p$.

Case 1: $\text{st}\in\{1,2,\infty\}$:

Let the adversarial examples for parameter $\theta_1$ and $\theta_2$ be
\[\bx_1=\arg\max_{\|\delta\|\leq\epsilon_p} g(\bx+\delta,\theta_1)\] \[\bx_2=\arg\max_{\|\delta\|\leq\epsilon_p} g(\bx+\delta,\theta_2),\]
then we have
	\begin{equation*}
\begin{aligned}
&\|h^{st}(\theta_1,z)-h^{st}(\theta_1,z)\|\\
=&|g(\theta_1,z_1)-g(\theta_2,z_2)|\\
\leq &\max\{|g(\theta_1,z_1)-g(\theta_2,z_1)|,|g(\theta_1,z_2)-g(\theta_2,z_2)|\}\\
\leq &L_{\theta}\|\theta_1-\theta_2\|,
\end{aligned}
\end{equation*}
where the first inequality is based on the fact that $g(\theta_1,z_1)\geq g(\theta_1,z_2)$ and $g(\theta_2,z_2)\geq g(\theta_2,z_1)$, the second inequality is based on Assumption 1\ref{ass2}. This proves the first inequality in equation (\ref{gst}) in this case. For the second one in equation (\ref{gst}), we have 
\begin{equation*}
\begin{aligned}
&\|\nabla_\theta h^{st}(\theta_1,z)-\nabla_\theta h^{st}(\theta_2,z)\|\\
=&\|\nabla_\theta h^{st}(\theta_1,z_1)-\nabla_\theta g(\theta_2,z_2)\|\\
\leq&\|\nabla_\theta h^{st}(\theta_1,z_1)-\nabla_\theta g(\theta_2,z_1)\|+\|\nabla_\theta g(\theta_2,z_1)-\nabla_\theta g(\theta_2,z_2)\|\\
\leq &  L_{\theta\theta}\|\theta_1-\theta_2\|+L_{\theta\bx}^p\|z_1-z_2\|_p\\
\leq &  L_{\theta\theta}\|\theta_1-\theta_2\|+L_{\theta\bx}^p[\|z_1-z\|_p+\|z-z_2\|_p]\\
\leq &  L_{\theta\theta}\|\theta_1-\theta_2\|+2L_{\theta\bx}^p\epsilon_p\\
= & L_{\theta\theta}\|\theta_1-\theta_2\|+\eta_{st},
\end{aligned}
\end{equation*}
where the first and the third inequality is triangle inequality, the second inequality is based on Assumption 1. This proves the second inequality.

Case 2: $\text{st}=\textbf{wst}$:

Let the adversarial examples for parameter $\theta_1$ and $\theta_2$ be
\[\bx_1=\arg\max_{p\in\{1,2,\infty\}}\max_{\|\delta\|\leq\epsilon_p} g(\bx+\delta,\theta_1)\] \[\bx_2=\arg\max_{p\in\{1,2,\infty\}}\max_{\|\delta\|\leq\epsilon_p} g(\bx+\delta,\theta_2),\]
the prove of the first inequality  in equation (\ref{gst}) is the same as the proof in Case 1. For the second inequality in equation (\ref{gst}), we have 
\begin{equation*}
\begin{aligned}
&\|\nabla_\theta h^{st}(\theta_1,z)-\nabla_\theta h^{st}(\theta_2,z)\|\\
\leq &  L_{\theta\theta}\|\theta_1-\theta_2\|+L_{\theta\bx}^p[\|z_1-z\|_p+\|z-z_2\|_p]\\
\leq &  L_{\theta\theta}\|\theta_1-\theta_2\|+2\max_{p\in\{1,2,\infty\}}[L_{\theta\bx}^p\epsilon_p]\\
= & L_{\theta\theta}\|\theta_1-\theta_2\|+\eta_{st}.
\end{aligned}
\end{equation*}
This proves the second inequality.

Case 3: $\text{st}=\textbf{avg}$:

Let the adversarial examples for parameter $\theta_1$ and $\theta_2$ and $p=1,2,\infty$ be
\[\bx_1^p=\arg\max_{\|\delta\|\leq\epsilon_p} g(\bx+\delta,\theta_1)\] \[\bx_2^p=\arg\max_{\|\delta\|\leq\epsilon_p} g(\bx+\delta,\theta_2).\]

For the first inequality in equation (\ref{gst}), we have
\begin{equation*}
\begin{aligned}
&\|h^{st}(\theta_1,z)-h^{st}(\theta_1,z)\|\\
=&|\mathbb{E}_{p\sim\{1,2,\infty\}}g(\bx_1^p,\theta_1)-\mathbb{E}_{p\sim\{1,2,\infty\}}g(\bx_2^p,\theta_2)|\\
\leq &\mathbb{E}_{p\sim\{1,2,\infty\}}|g(\bx_1^p,\theta_1)-g(\bx_2^p,\theta_2)|\\
\leq &\mathbb{E}_{p\sim\{1,2,\infty\}}\max\{|g(\bx_1^p,\theta_1)-g(\bx_1^p,\theta_2)|,|g(\bx_2^p,\theta_1)-g(\bx_2^p,\theta_2)|\}\\
\leq &\mathbb{E}_{p\sim\{1,2,\infty\}}L_{\theta}\|\theta_1-\theta_2\|\\
\leq &L_{\theta}\|\theta_1-\theta_2\|,
\end{aligned}
\end{equation*}
where the first inequality is Jensen's inequality. This proves of the first inequality in equation (\ref{gst}) in this case. For the second inequality in equation (\ref{gst}), we have 
\begin{equation*}
\begin{aligned}
&\|\nabla_\theta h^{st}(\theta_1,z)-\nabla_\theta h^{st}(\theta_2,z)\|\\
= &\|\nabla_\theta\mathbb{E}_{p\sim\{1,2,\infty\}}g(\bx_1^p,\theta_1)-\nabla_\theta \mathbb{E}_{p\sim\{1,2,\infty\}}g(\bx_2^p,\theta_2)\|\\
\leq& \mathbb{E}_{p\sim\{1,2,\infty\}} \|\nabla_\theta g(\bx_1^p,\theta_1)-\nabla_\theta g(\bx_2^p,\theta_2)\|\\
\leq&\mathbb{E}_{p\sim\{1,2,\infty\}} [L_{\theta\theta}\|\theta_1-\theta_2\|+2L_{\theta\bx}^p\epsilon_p]\\
= & L_{\theta\theta}\|\theta_1-\theta_2\|+\eta_{st},
\end{aligned}
\end{equation*}
where the first inequality is the Jensen's inequality, the second one is the result in Case 1. the This proves the second inequality in equation (\ref{gst}) in this case.\qed

\subsection{Proof of Lemma \ref{updaterules}}\label{proofl1}
\begin{equation*}
\begin{aligned}
\|G_z(\theta_1)-G_z(\theta_2)\|&= \|\theta_1-\theta_2-\frac{1}{P}\sum_{p=1}^P\alpha^p h^p(\theta_1,z)+\frac{1}{P}\sum_{p=1}^P\alpha^p h^p(\theta_2,z)\|\\
&\leq \frac{1}{P}\sum_{p=1}^P\|\theta_1-\theta_2-\alpha^p h^p(\theta_1,z)+\alpha^p h^p(\theta_2,z)\|\\
&\leq \frac{1}{P}\sum_{p=1}^P\|\theta_1-\theta_2\|\\
&= \|\theta_1-\theta_2\|,\\
\end{aligned}
\end{equation*}
where the first inequality is due to triangular inequality, the second inequality is due to the co-coercive propertiy of convex function.\qed
\subsection{Proof of Theorem \ref{gen}}\label{a4}
To bound the generalization gap of a model, we employ the
following notion of uniform stability.

\begin{definition}
A randomized algorithm $A$ is $\varepsilon$-\emph{uniformly stable} if
for all data sets $S,S'\in \mathcal{Z}^n$ such that $S$ and $S'$ differ in at most one
example, we have
\begin{equation}\label{eq:stab}
\sup_{z} \E_{A} \left[ h(A(S); z) - h(A(S'); z) \right] \le \varepsilon\,.
\end{equation}
\end{definition}
Here, the expectation is taken over the randomness of $A$. Uniform stability implies generalization in expectation \citep{hardt2016train}.

\begin{theorem}[Generalization in expectation]
\label{thm:stab2gen}
Let $A$ be $\varepsilon$-uniformly stable. Then, the expected generalization gap satisfies
\[
|\mathcal{E}_{gen}|=\left| \E_{S,A}[R_{\cal D}[A(S)] - R_S[A(S)]]\right| \le \varepsilon\,.
\]
\end{theorem}
Let $S$ and $S'$ be two samples of size $n$ differing in only a single
example. Consider two trajectories $\theta_1^1,\dots,\theta_1^T$ and $\theta_2^1,\dots,\theta_2^T$
induced by running an algorithm on sample $S$ and $S',$ respectively. Let $\delta_t=\|\theta_1^t-\theta_2^t\|$.

Fixing an example $z\in Z$ and apply
the Lipschitz condition on $h(\cdot\,;z)$, we have
\begin{equation}\label{eq:convex-diff}
\E\left|h(\theta_1^T;z)-h(\theta_2^T;z)\right| \le
L\E\left[\delta_T\right]\,.
\end{equation}

Observe that at step $t,$ with probability $1-1/n,$ the
example selected by the randomized algorithms is the same in both $S$ and $S'.$ 
With probability $1/n$ the selected example is
different. Based on Lemma \ref{updaterules}, we have

\begin{align}\label{eq:convex-recursion}
\E\left[\delta_{t+1}\right]
 &\le \left(1-\frac1n\right)\bigg(\E[\delta_t]\bigg) +
 \frac1n\E\left[\delta_t\right]  +
\frac{2\frac{1}{P}\sum_{p=1}^P\alpha_t^p L}n \\
&\le\E\left[\delta_t\right] + \bigg(\eta+\frac{2L}{n}\bigg)\frac{1}{P}\sum_{p=1}^P\alpha_t^p\,.
\end{align}

Unraveling the recursion, we have
\[
\E\left[\delta_T\right] \le\bigg(\frac{2L}{n}\bigg)\sum_{t=1}^T\frac{1}{P}\sum_{p=1}^P\alpha_t^p,\ \text{and}\ \ 
\mathcal{E}_{gen} \le L \bigg(\frac{2L}{n}\bigg)\sum_{t=1}^T\frac{1}{P}\sum_{p=1}^P\alpha_t^p\,.
\]
Since this bounds holds for all $S,S'$ and $z,$ we obtain the desired bound on
the uniform stability.\qed
\subsection{Proof of Theorem \ref{opt}}\label{a5}
\begin{eqnarray*}
&&\|\theta^{t+1}-\theta^*\|^2
= \bigg\|\theta^{t}-\theta^*-\frac{1}{P}\sum_{p=1}^P\alpha_p^t \nabla h^p(\theta^t,z)\bigg\|^2\\
&=& \|\theta^{t}-\theta^*\|^2+\bigg\|\frac{1}{P}\sum_{p=1}^P\alpha_p^t \nabla h^p(\theta^t,z)\bigg\|^2-2\bigg\langle \frac{1}{P}\sum_{p=1}^P\alpha_p^t \nabla h^p(\theta^t,z), \theta^{t+1}-\theta^*\bigg\rangle.
\end{eqnarray*}
Take expectation over $z$, we have
\begin{eqnarray*}
&&\mathbb{E}\|\theta^{t+1}-\theta^*\|^2\\
&\le& \mathbb{E}\|\theta^{t}-\theta^*\|^2 + \bigg(\frac{1}{P}\sum_{p=1}^P\alpha_p^t L\bigg)^2 - \frac{2}{P} \sum_{p=1}^P\alpha_p^t \mathbb{E}[h^p(\theta^t)-h^p(\theta^*)],
\end{eqnarray*}
Then, 
\begin{eqnarray*}
&&\alpha_{sw}^t\frac{2}{P} \sum_{p=1}^P \mathbb{E}[h^p(\theta^t)-h^p(\theta^*)]
\leq \mathbb{E}\|\theta^{t}-\theta^*\|^2-\mathbb{E}\|\theta^{t+1}-\theta^*\|^2\\
&&+(\alpha_{sw}^t L)^2 + \frac{2}{P} \sum_{p=1}^P(\alpha_{sw}^t-\alpha_p^t) \mathbb{E}[h^p(\theta^t)-h^p(\theta^*)]\\
\end{eqnarray*}
Considering constant step size and expand the recursive, we have
\begin{eqnarray*}
&&T\alpha_{sw}\frac{2}{P} \sum_{p=1}^P \mathbb{E}[h^p(\theta^T)-h^p(\theta^*)]\\
&\leq& \mathbb{E}\|\theta^{T}-\theta^*\|^2+ T(\alpha_{sw} L)^2 + \frac{2T}{P} \sum_{p=1}^P|\alpha_{sw}^t-\alpha_p^t|B.\\
\end{eqnarray*}
Therefore, we obtain the optimization error bound
\[
\mathcal{E}_{opt}\leq \frac{D^2+L^2T\alpha_{sw}^2}{2T\alpha_{sw}}+B\frac{\sum_{p=1}^P|\alpha_{sw}-\alpha_p|}{\alpha_{sw}}.
\]\qed

\section{Additional Experiments}
\label{C}
In this section, the test accuracy are evaluated using the first batch (128 samples) to save the computational cost.
\subsection{Gradient Norm Analysis}
\begin{figure*}[t]
\centering
\scalebox{0.9}{
\subfigure[]{
\begin{minipage}[htp]{0.24\linewidth}
\centering
\includegraphics[width=1.2in]{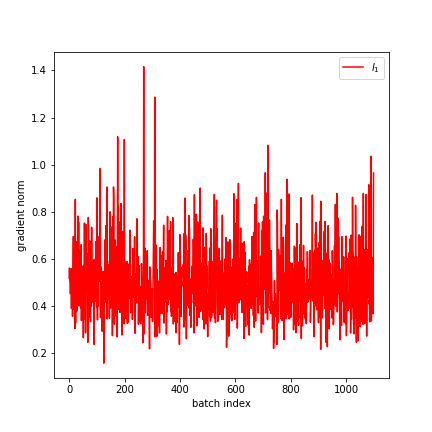}
\end{minipage}%
}
\subfigure[]{
\begin{minipage}[htp]{0.24\linewidth}
\centering
\includegraphics[width=1.2in]{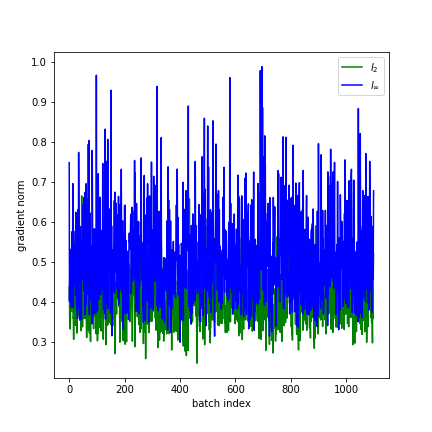}
\end{minipage}
}
\subfigure[]{
\begin{minipage}[htp]{0.24\linewidth}
\centering
\includegraphics[width=1.2in]{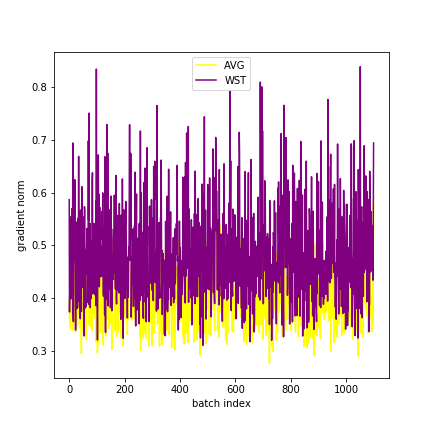}
\end{minipage}
}
\subfigure[]{
\begin{minipage}[htp]{0.24\linewidth}
\centering
\includegraphics[width=1.2in]{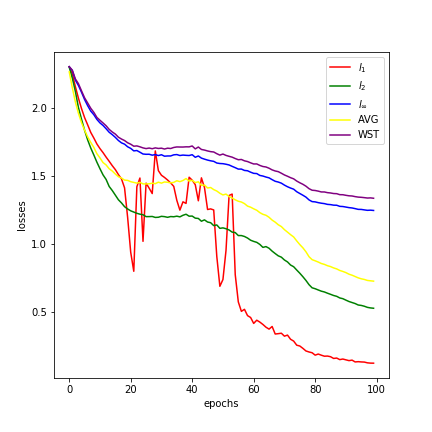}
\end{minipage}
}}
\centering
\vskip -0.15in
\caption{Gradient norms and loss value of adversarial training for single and multiple perturbations. (a) $\|\nabla R_{S}^1(\theta)\|$, last 1173 batch. (b) $\|\nabla R_{S}^2(\theta)\|$ (green) and $\|\nabla R_{S}^\infty(\theta)\|$ (blue), last 1173 batch. (c) $\|\nabla R_{S}^{avg}(\theta)\|$ (yellow) and $\|\nabla R_{S}^{wst}(\theta)\|$ (purple), last 1173 batch. (d) Training loss for the total 100 epochs.}
\label{fig:gradnorm}
\vskip -0.1in
\end{figure*}

Since the gradient Lipschitz $L_{\theta x}^p$ is unknown in practice, we provide a numerical simulation of the convergence error in this subsection on CIFAR-10 to help justify our Theoretical results. In Fig. \ref{fig:gradnorm}, we show the gradient norm $\|\nabla R_{S}^{st}(\theta)\|$ of the last layer for the last 3 epochs ($3\times391=1173$ batchs) for $st\in\{1,2,\infty,avg,wst\}$ as well as the training loss\footnote{We should use optimality gap (training loss - optimal loss) to evaluate convergence, but the optimal loss is unknown, we use training loss as a substitude.} for the total 100 epochs.
\paragraph{Comparison of $\ell_1$, $\ell_2$, and $\ell_\infty$} In Fig. \ref{fig:gradnorm}, We can see that $\|\nabla R_{S}^1(\theta)\|$ is the largest one accompany with the largest variance among these three. In (d), the training loss of $ R_{S}^1(\theta)$ is unstable. It is because the top-k $\ell_1$ attack is inefficient and sparse \citep{tramer2019adversarial}. In the middle stage, the fluctuation is large since the success rate of the top-k attack is small. In the final stage, the top-k attack cannot find adversarial examples. Therefore, the loss is small.  Comparing the $ R_{S}^2(\theta)$ and $ R_{S}^\infty(\theta)$, we can see that $\ell_\infty$ adversarial training has higher gradient norm and larger training loss. In conclusion, $\ell_1$ and $\ell_\infty$ give less contributions to the smoothness of ATMP.

\subsection{Ablation Study of SWA}
\begin{figure}[t]
\centering
\scalebox{0.9}{
\subfigure[]{
\begin{minipage}[htp]{0.24\linewidth}
\centering
\includegraphics[width=1.2in]{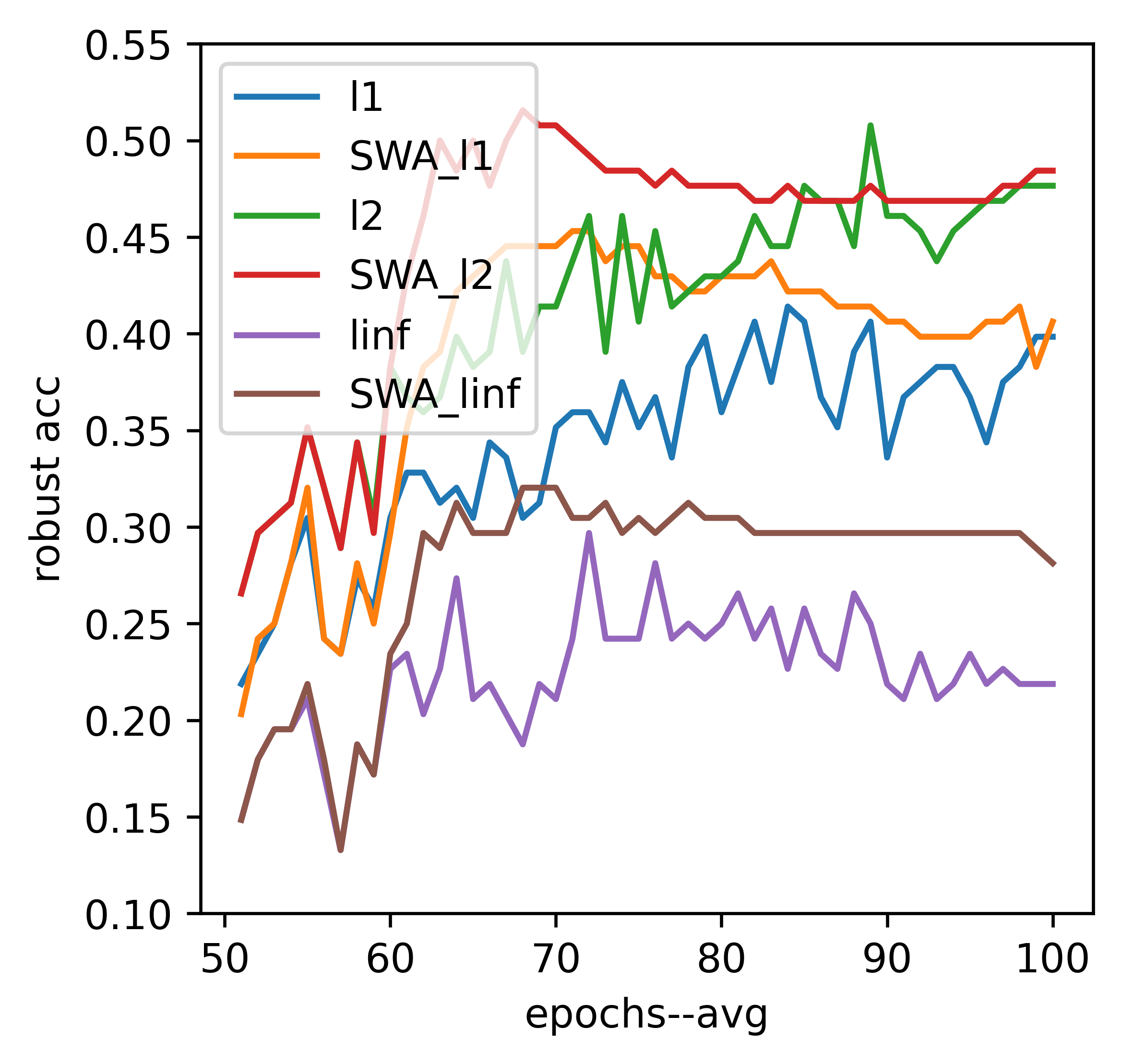}
\end{minipage}%
}
\subfigure[]{
\begin{minipage}[htp]{0.24\linewidth}
\centering
\includegraphics[width=1.2in]{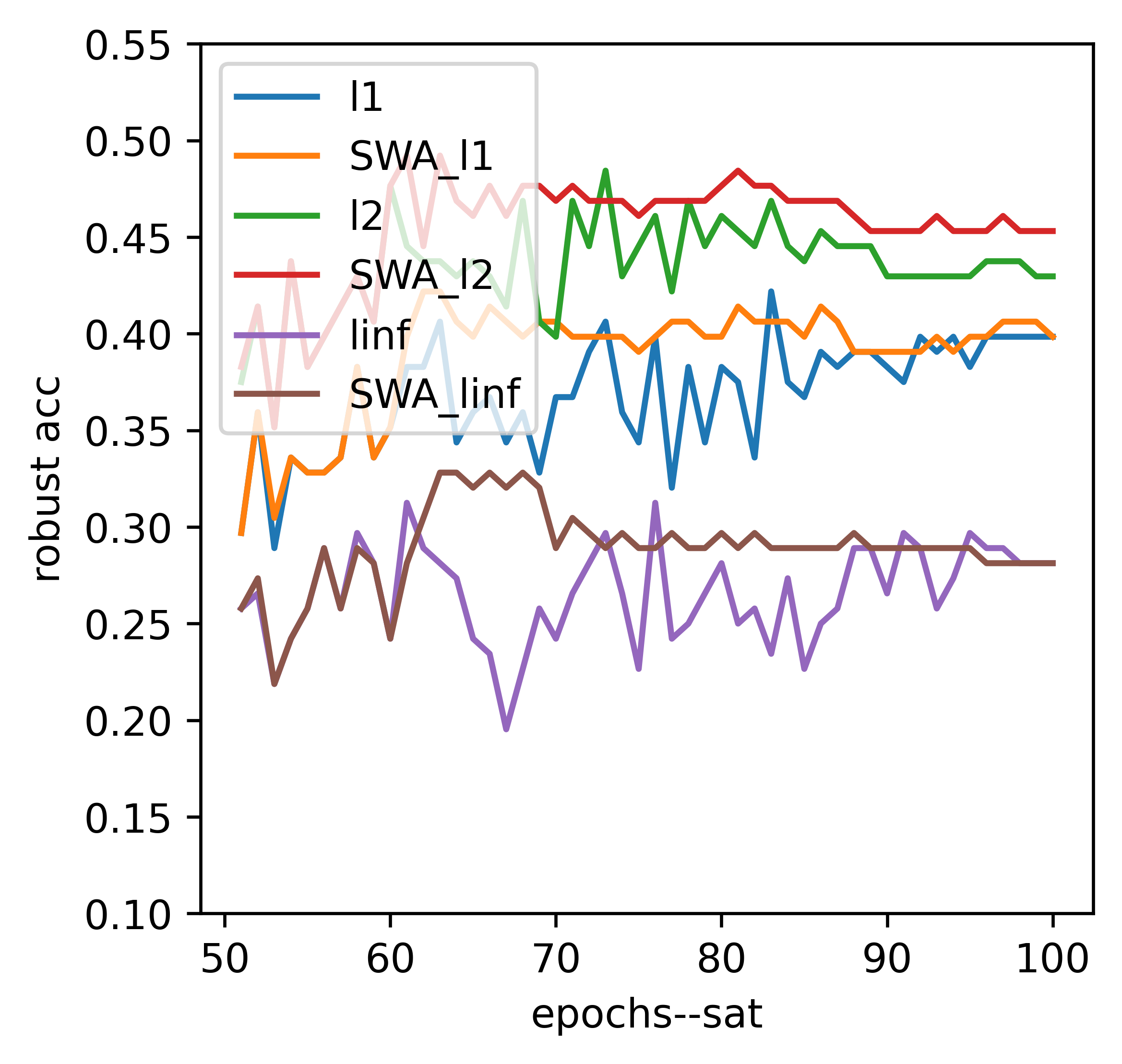}
\end{minipage}
}
\subfigure[]{
\begin{minipage}[htp]{0.24\linewidth}
\centering
\includegraphics[width=1.2in]{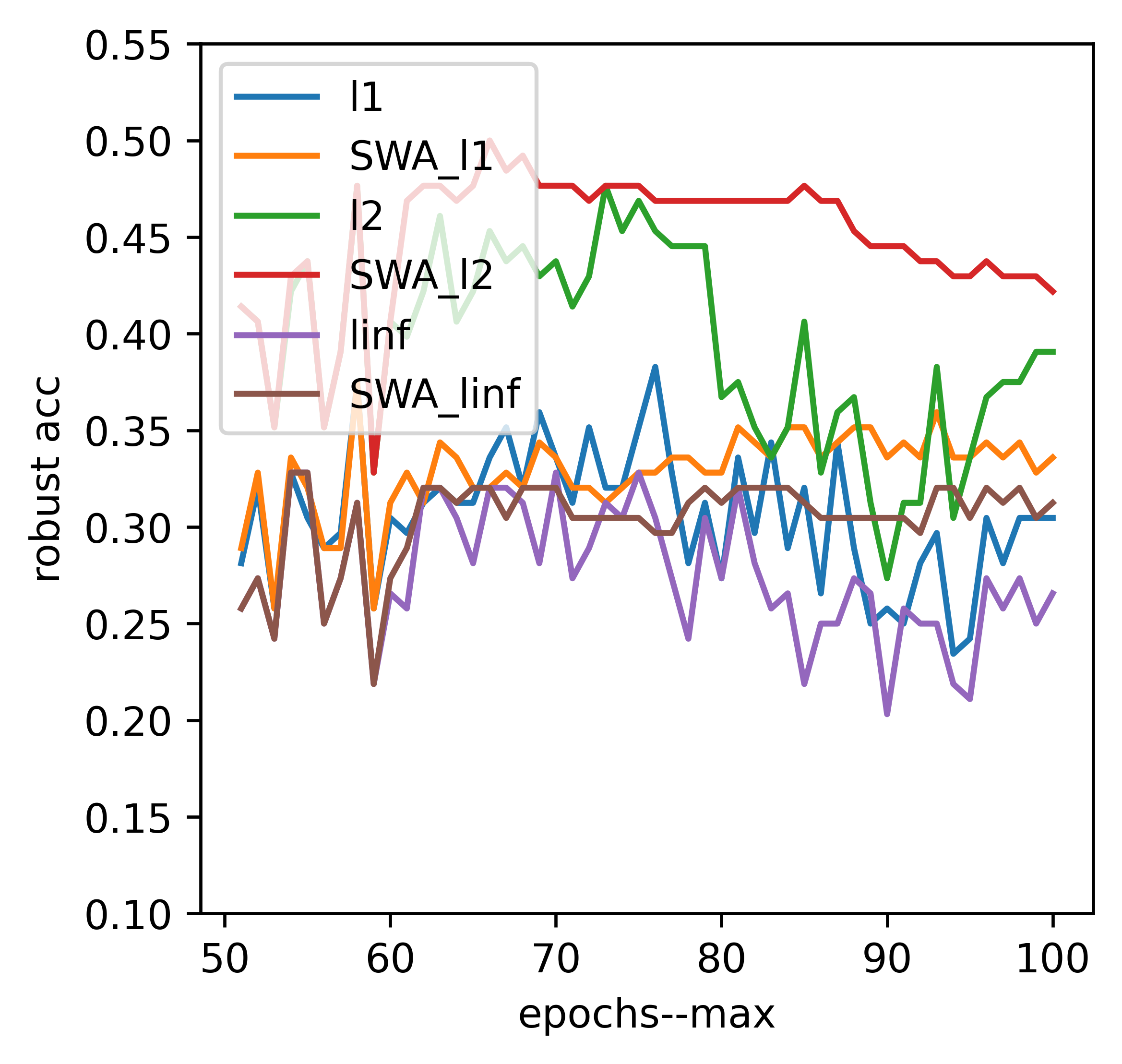}
\end{minipage}
}
\subfigure[]{
\begin{minipage}[htp]{0.24\linewidth}
\centering
\includegraphics[width=1.2in]{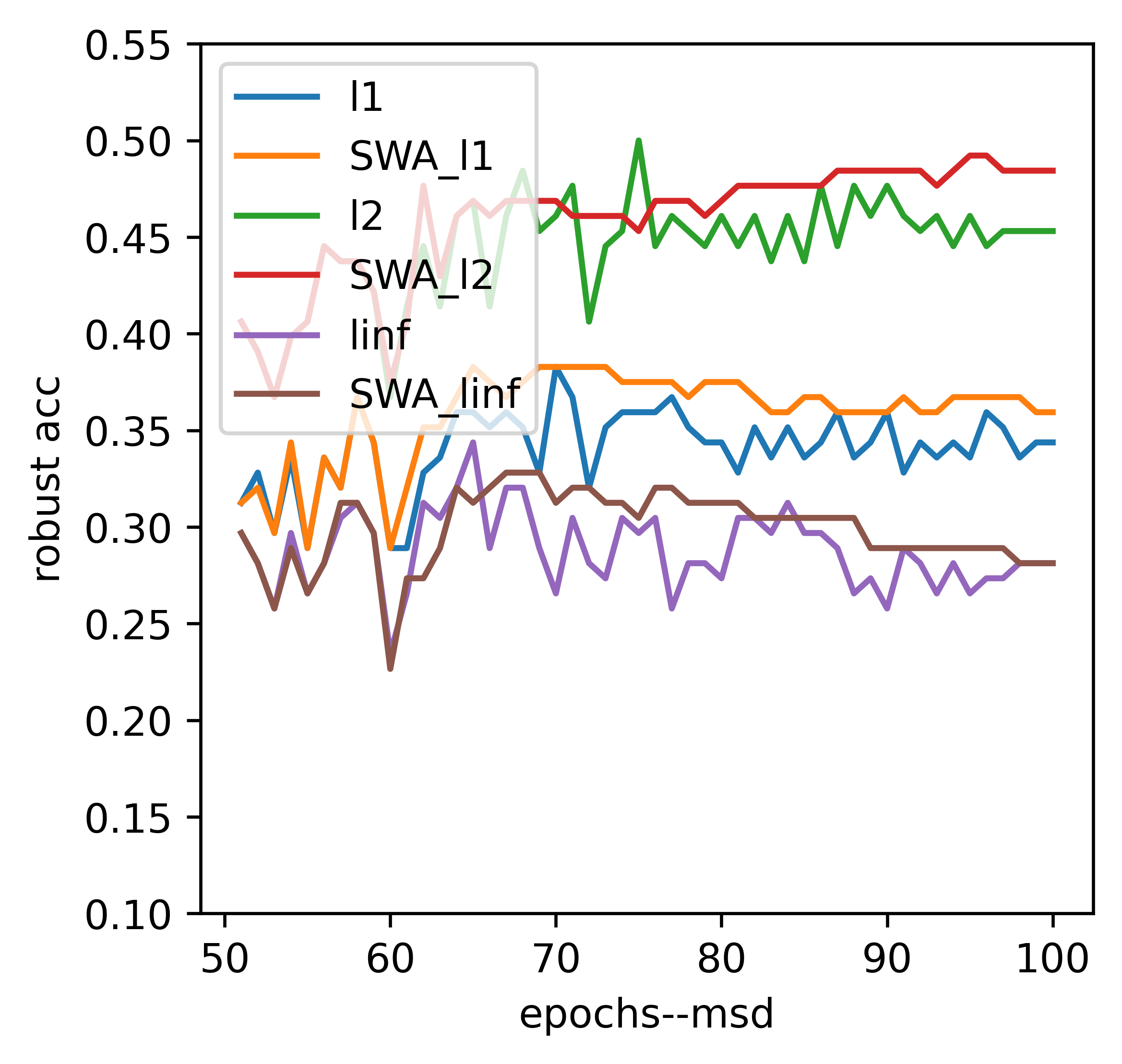}
\end{minipage}
}}
\centering
\vskip -0.15in
\caption{Results of robust accuracy for each type perturbation over epochs  of multiple perturbation AT in CIFAR-100 dataset:(a) Avg and Avg-SWA, (b) SAT and SAT-SWA, (c) Max and Max-SWA, (d) MSD and MSD-SWA.}
\label{cifar100}
\vskip -0.1in
\end{figure}

We give more ablation study of SWA in this subsection.
\paragraph{SWA on CIFAR-100} In Fig. \ref{cifar100}, we we plots the robust accuracy of of ATMP using AVG, SAT, MAX, and MSD respectively on CIFAR-100, which are the same experiments we show in Figure 3. From the plots, we observe that without SWA, the test accuracy are highly unstable among different training epochs. There exists a trade-off between different types of adversaries. The increase in robust accuracy against one kind of adversaries may accompany with the decrease in robust accuracy on another adversarial examples. On the other side, when coupling with SWA, the tendency curves of all four ATMP strategies are largely stabilized. Besides, SWA (red, orange, and brown line) will increase the test accuracy in most of the case. The experiments in Figure \ref{cifar100} give the same conclusion as that of Figure \ref{cifar10} in the main paper.

\begin{figure}[t]
\centering
\scalebox{0.9}{
\subfigure[]{
\begin{minipage}[htp]{0.40\linewidth}
\centering
\includegraphics[width=2in]{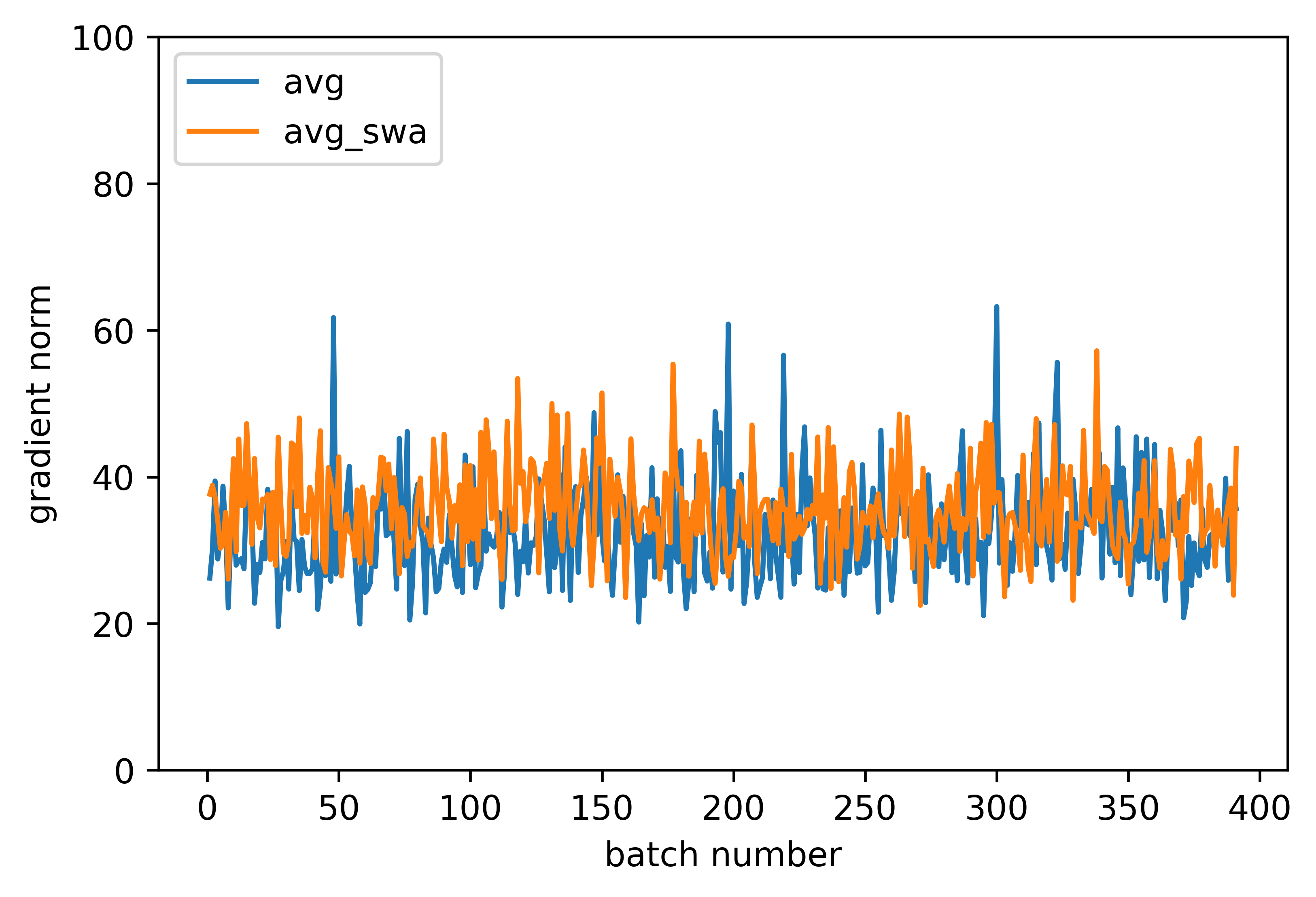}
\end{minipage}%
}
\subfigure[]{
\begin{minipage}[htp]{0.60\linewidth}
\centering
\includegraphics[width=2in]{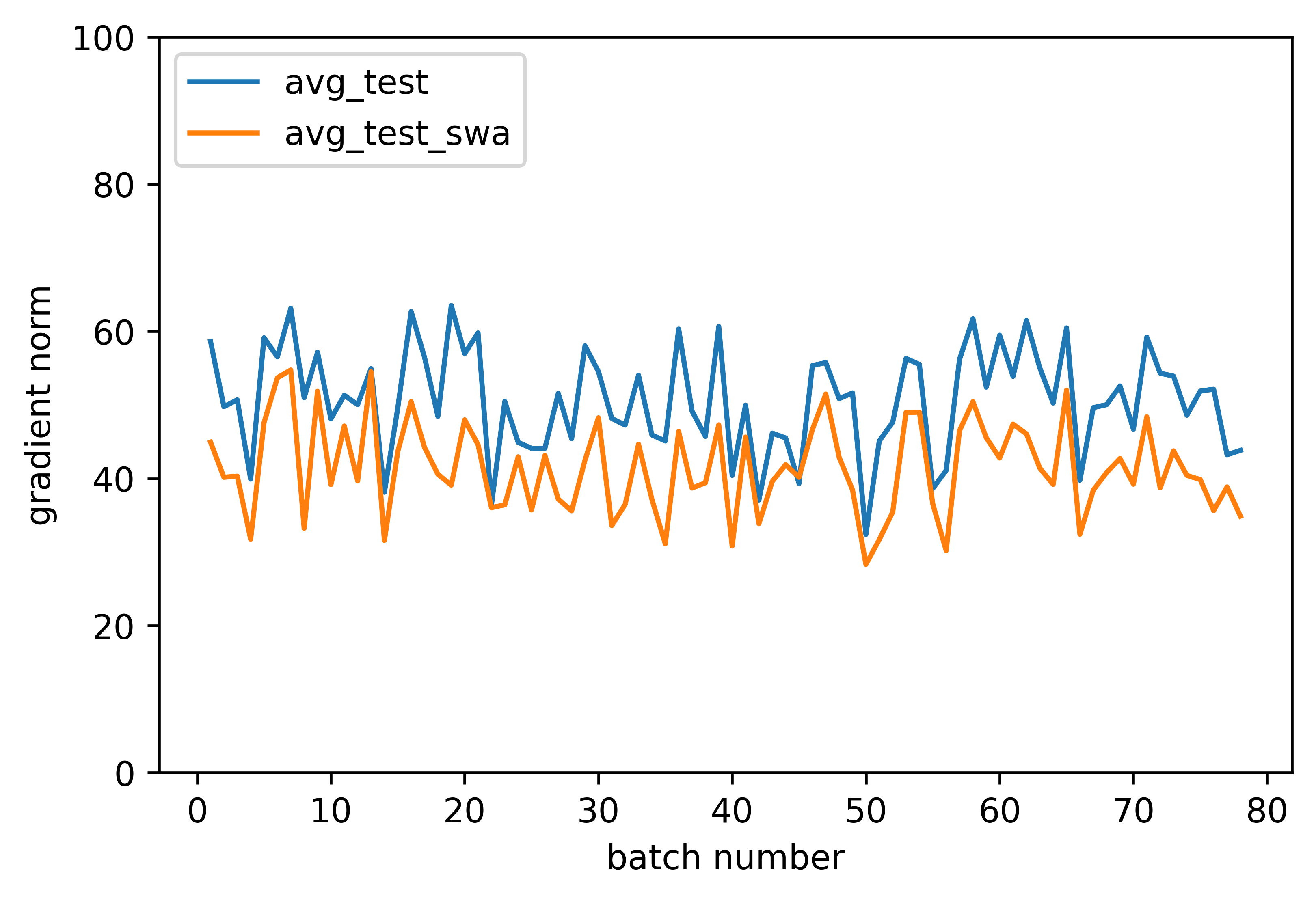}
\end{minipage}
}}
\scalebox{0.9}{
\subfigure[]{
\begin{minipage}[htp]{0.40\linewidth}
\centering
\includegraphics[width=2in]{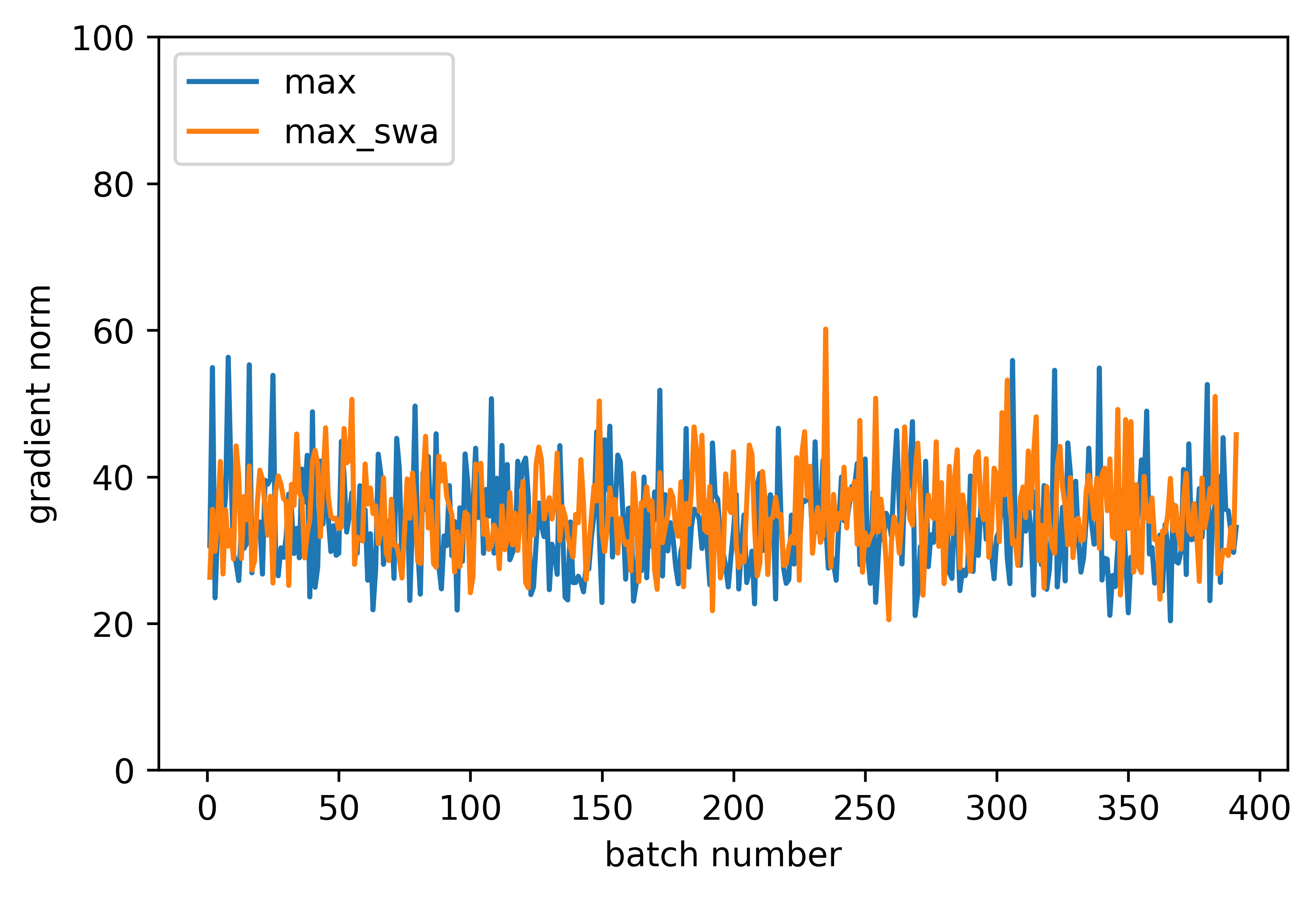}
\end{minipage}%
}
\subfigure[]{
\begin{minipage}[htp]{0.60\linewidth}
\centering
\includegraphics[width=2in]{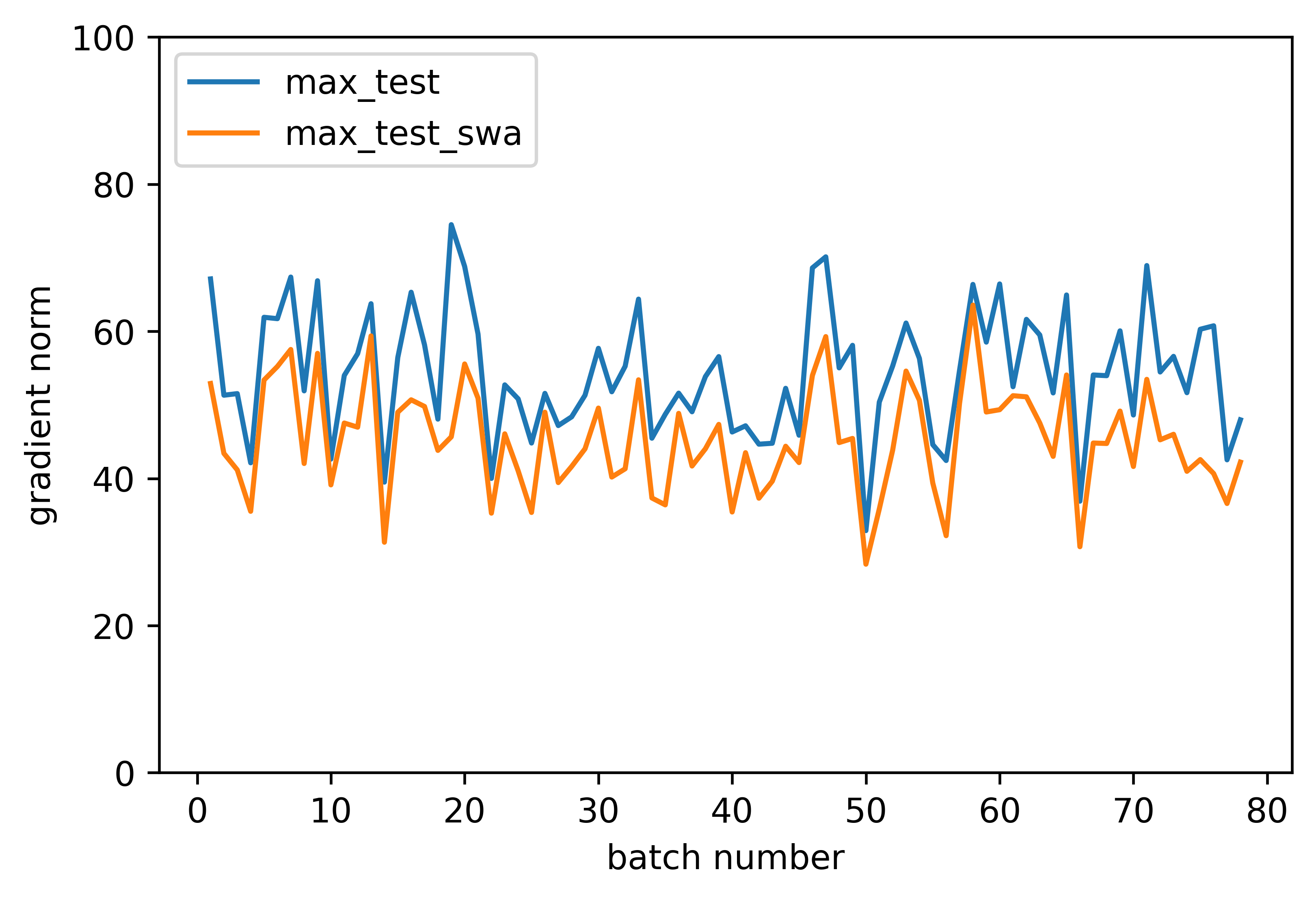}
\end{minipage}
}}
\centering
\vskip -0.15in
\caption{Gradient norm with and without SWA.}
\label{gradnormswa}
\vskip -0.1in
\end{figure}

\paragraph{Gradient norm of SWA} In Figure \ref{gradnormswa}, we show the gradient norm of all the batches with and without SWA using AVG and MAX. On training set, the gradient norm with and without SWA is similar. On test set, the gradient norm of using SWA is smaller than that without SWA. This proves that SWA can find flatter minima with smaller gradient norm, and have better generalization.


\paragraph{When to start SWA} In Table \ref{SWAepoch} we try to start SWA at the 60th, 70th, and 80th epochs. We use MSD for our experiments. We can see that the test accuracy against $\ell_\infty$ attack is higher when we start later. BUt the test accuracy against $\ell_1$ attack is higher when we start earlier. It is hard to say which is better but using SWA is always better than no SWA. The accuracy of AVG, MAX MSD, and SAT with and without SWA are provided in Table \ref{table1}.

\begin{table}[htbp]
    \centering
    \caption{Test accuracy (\%) of MSD with SWA started from the 60th,70th, and 80th epochs.}
    \begin{tabular}{c|c|cccc}
    \hline
      \multicolumn{2}{c|}{Dataset}&\multicolumn{4}{c}{CIFAR-10}\\ \hline
         \multicolumn{2}{c|}{Attack methods}& clean & $\ell_1$ & $\ell_2$ & $\ell_\infty$  \\ \hline
        \multirow{3}{*}{MSD with SWA}& no SWA & 83.51 & 50.78 & 66.41 & 43.75 \\ 
        &the 60th epochs & 81.99 & 53.12 & 68.75 & 46.09 \\
        & the 70th epochs& 82.01 & 49.22 & 69.53 & 47.66\\
        & the 80th epochs & 81.95 & 50.78 & 69.54 & 48.44\\
        \hline

    \end{tabular}
    \label{SWAepoch}
\end{table}

\begin{table*}
    \centering
    \caption{Test accuracy (\%) of different algorithms (MAX, AVG, MSD and SAT, with and without SWA) against $\ell_1$, $\ell_2$, and $\ell_\infty$ attacks on CIFAR-10 and CIFAR-100.}
    \resizebox{\linewidth}{!}{%
    \begin{tabular}{c|c||cccc||cccc}
    \hline
      \multicolumn{2}{c||}{Dataset}&\multicolumn{4}{c||}{CIFAR-10}&\multicolumn{4}{|c}{CIFAR-100}\\ \hline
         \multicolumn{2}{c||}{Attack methods}& clean & $\ell_1$ & $\ell_2$ & $\ell_\infty$ & clean & $\ell_1$  & $\ell_2$ & $\ell_\infty$  \\ \hline
        \multirow{3}{*}{AT}&$\ell_1$ & 93.19 & 89.84 & 0.00 & 0.00 &  70.98 & 73.44 & 00.78 & 00.78 \\ 
        &$\ell_2$ & 88.66 & 41.41 & 61.72 & 39.84 &  63.76 & 21.88 & 43.75 & 20.31 \\
        &$\ell_\infty$ & 84.94 & 17.19 & 53.91 & 46.88  & 58.86 & 11.72 & 39.06 & 30.47 \\ \hline
        \multirow{4}{*}{ATMP}&AVG & \textbf{85.28} & 50.78 & 64.84 & 38.28 &  59.71 & 39.84 & 47.66 & 21.88\\
        &MAX & 84.96 & 43.75 & 49.22 & 41.41 & 57.90 & 30.47 & 39.06 & 26.56 \\ 
        &MSD & 83.51 & 50.78 & 66.41 & 43.75 &  57.33 & 34.38 & 45.31 & 28.12 \\ 
        &SAT & 85.23 & 53.12 & 69.53 & 40.62 &  59.25 & 39.84 & 42.97 & 28.12 \\ \hline
        \multirow{4}{0.6in}{\minitab[c]{ATMP\\ with\\ SWA}}&AVG & 83.19 & 57.03 & 69.53 & 42.97 &  58.04 & \textbf{40.62} & \textbf{48.44} & 28.12 \\ 
        &MAX & 83.78 & 42.19 & 60.16 & \textbf{47.66} &  58.29 & 33.59 & 42.19 & \textbf{31.25} \\ 
        &MSD & 81.99 & 53.12 & 68.75 & 46.09 &  57.33 & 35.94 & \textbf{48.44} & 28.12 \\ 
        &SAT & 84.76 & \textbf{59.38} & \textbf{71.09} & 42.97  & \textbf{60.18} & 39.84 & 45.31 & 28.12 \\ 
        \hline
    \end{tabular}}
    \label{table1}
\end{table*}

\subsection{Other Tricks}

\paragraph{Label smoothing (LS)} Label smoothing is introduced as regularization to improve generalization
by replacing the one-hot labels to soft labels in the cross-entropy loss. It is shown that it relates to flat minima and yields better generalization. In the cross-entropy loss, using soft labels other than one-hot label can increase the smoothness of the loss function. Given probability $\gamma\in [0,1]$ and number of classes $K$, for each sample $(\bx_i,y_i)$, let $y_i$ keeps the true label with probability $1-\gamma$, $y_i$ is replace by a wrong label in one of the other $K-1$ classes with equal probability $\gamma/(K-1)$. The performance of ATMP with label smoothing is provided in Table \ref{table_ls}. We can see that label smoothing provides some improvement in these four strategies.

\label{b3}
\begin{table*}
    \centering
    \caption{Test accuracy (\%) of different algorithms (MAX, AVG, MSD and SAT, with label smoothing) against $\ell_1$, $\ell_2$, and $\ell_\infty$ attacks on CIFAR-10 and CIFAR-100.}
    \resizebox{\linewidth}{!}{%
    \begin{tabular}{c|c||cccc||cccc}
    \hline
          \multicolumn{2}{c||}{Dataset}&\multicolumn{4}{c||}{CIFAR-10}&\multicolumn{4}{|c}{CIFAR-100}\\ \hline
         \multicolumn{2}{c||}{Attack methods}& clean & $\ell_1$ & $\ell_2$ & $\ell_\infty$ & clean & $\ell_1$  & $\ell_2$ & $\ell_\infty$  \\ \hline
         \multirow{4}{*}{\minitab[c]{ATMP\\ with\\LS}}&AVG&\textbf{85.57} & 55.47 & 68.75 & 39.06 & \textbf{60.58} & 38.28 & 46.88 & 24.22 \\ 
         &MAX & 84.49 & 45.31 & 57.03 & 45.31 & 58.25 & 33.59 & 45.31 & 29.69 \\ 
         &MSD & 83.49 & 50.78 & 69.53 & 45.30 & 58.61 & 35.94 & \textbf{48.44} & 32.03 \\ 
        &SAT  &85.70 & 53.91 & 67.97 & 37.50 & 60.17 & 39.84 & 44.53 & 26.56 \\ \hline
         \multirow{4}{0.6in}{\minitab[c]{ATMP\\ with\\LS \\ \& SWA}}&AVG &83.42 & \textbf{57.03} & 71.09 & \textbf{48.44} & 59.09 & \textbf{42.19} & 45.31 & 28.12 \\ 
         &MAX &83.05 & 47.66 & 63.28 & 46.09 & 57.76 & 39.84 & 47.66 & 31.25 \\ 
         &MSD &82.15 & 50.78 & \textbf{71.88} & 46.09 & 57.70 & 38.28 & 46.88 &\textbf{35.94} \\ 
        &SAT &84.96 & 56.25 & 66.41 & 41.41 & 60.51 & 39.06 & 47.66 & 29.69 \\ \hline
    \end{tabular}}
    \label{table_ls}
\end{table*}

\paragraph{Label noise} Label noise is a alternative choose of label smoothing. Given probability $\gamma\in [0,1]$ and number of classes $K$, for each sample $(\bx_i,y_i)$, let $y_i$ keeps the true label with probability $1-\gamma$, $y_i$ is replace by a wrong label in one of the other $K-1$ classes with equal probability $\gamma/(K-1)$. The difference between label noise and label smoothing is that label noise use hard label while label smoothing use soft label.

In Table \ref{ln}, we provide the experiments of incorporating label noise in ATMP. We can see that label noise can give some improvements. But the improvements are not comparable to label smoothing. The reason behind is that label noise cannot increase the smoothness of the loss function of ATMP.
\begin{table}[htbp]
    \centering
    \caption{Test accuracy (\%) of ATMP with label noise.}
    \begin{tabular}{c|c|cccc}
    \hline
      \multicolumn{2}{c|}{Dataset}&\multicolumn{4}{c}{CIFAR-10}\\ \hline
         \multicolumn{2}{c|}{Attack methods}& clean & $\ell_1$ & $\ell_2$ & $\ell_\infty$  \\ \hline
        \multirow{4}{*}{ATMP with LN}& MAX & 84.18 & 43.75 & 57.03 & 44.53 \\ 
        & AVG & 84.67 & 57.03 & 70.31 & 41.41 \\
        & MSD & 82.79 & 52.34 & 6562 & 47.66\\
        & SAT & 85.36 &56.25 &69.53 &42.19\\
        \hline
        \multirow{4}{*}{ATMP with LN \& SWA}& MAX & 81.66 & 50.78 & 63.28 & 47.66 \\ 
        & AVG & 81.06& 57.81 & 70.31 & 42.97 \\
        & MSD & 81.03  & 51.56 & 64.84 & 50.00\\
        & SAT & 84.00 &59.38& 67.19& 46.88\\
         \hline
    \end{tabular}
    \label{ln}
\end{table}

\paragraph{Silu activation function} Silu activation function is proposed to deal with the non-smoothness of Relu activation function. In Table \ref{silu}, we can see that Silu has some small improvement on ATMP.

\begin{table}[htbp]
    \centering
    \caption{Test accuracy (\%) of ATMP with Silu activation function.}
    \begin{tabular}{c|c|cccc}
    \hline
      \multicolumn{2}{c|}{Dataset}&\multicolumn{4}{c}{CIFAR-10}\\ \hline
         \multicolumn{2}{c|}{Attack methods}& clean & $\ell_1$ & $\ell_2$ & $\ell_\infty$  \\ \hline
        \multirow{4}{*}{ATMP with Silu}& MAX & 80.91 & 53.91 & 64.84 & 46.09 \\ 
        & AVG & 81.61 & 57.03 & 70.31 & 42.19 \\
        & MSD & 80.44 & 57.03 & 64.84 & 48.44\\
        & SAT & 83.89& 54.69&64.84&44.53\\
        \hline
        \multirow{4}{*}{ATMP with Silu \& SWA}& MAX & 78.48 & 50.78 & 67.19 & 49.22 \\ 
        & AVG & 78.20& 58.59 & 68.75 & 42.98 \\
        & MSD & 78.39 & 55.47 & 66.41 & 49.99\\
        & SAT & 82.02 &57.81&67.19&45.31\\
         \hline
    \end{tabular}
    \label{silu}
\end{table}
\paragraph{Mixup} We use adversarial examples in the form $\bx_{adv}=\bx+(\delta_1+\delta_2+\delta_\infty)/3$ for adversarial training. In Table \ref{mixup}, we can see that the robust accuracy is not comparable to MAX and AVG. It is because this training procedure focuses more on the mixup adversarial examples. It fails to fit the three types of adversarial examples.

\begin{table}[htbp]
    \centering
    \caption{Test accuracy (\%) of ATMP using mixup.}
    \begin{tabular}{c|c|cccc}
    \hline
      \multicolumn{2}{c|}{Dataset}&\multicolumn{4}{c}{CIFAR-10}\\ \hline
         \multicolumn{2}{c|}{Attack methods}& clean & $\ell_1$ & $\ell_2$ & $\ell_\infty$  \\ \hline
        Other tricks & Mixup & 84.73 & 51.56 & 67.19 & 39.06 \\
        \hline
    \end{tabular}
    \label{mixup}
\end{table}

\end{document}